\documentclass[9pt,lineno]{elife}

\usepackage{lipsum} 
\usepackage[version=4]{mhchem}
\usepackage{siunitx}
\usepackage{bm} 
\usepackage{multirow}
\DeclareSIUnit\Molar{M}

\title{Information-Restricted Neural Language Models Reveal Different Brain Regions' Sensitivity to Semantics, Syntax and Context}

\author[1, 2]{Alexandre Pasquiou}
\author[1]{Yair Lakretz}
\author[2]{Bertrand Thirion}
\author[1]{Christophe Pallier}
\affil[1]{UNICOG, Cognitive Neuroimaging Unit, INSERM, CEA, Neurospin, Gif-sur-Yvette, France}
\affil[2]{MIND, INRIA, CEA, Neurospin, Gif-sur-Yvette, France}

\corr{alexandre.pasquiou@inria.fr}{FMS}

\begin{document}
\nolinenumbers
\maketitle

\begin{abstract}
A fundamental question in neurolinguistics concerns the brain regions involved in syntactic and semantic processing during speech comprehension, both at the lexical (word processing) and supra-lexical levels (sentence and discourse processing).
To what extent are these regions separated or intertwined?
To address this question, we trained a lexical language model, Glove, and a supra-lexical language model, GPT-2, on a text corpus from which we selectively removed either syntactic or semantic information.
We then assessed to what extent these information-restricted models were able to predict the time-courses of fMRI signal of humans listening to naturalistic text.
We also manipulated the size of contextual information provided to GPT-2 in order to determine the windows of integration of brain regions involved in supra-lexical processing.
Our analyses show that, while most brain regions involved in language are sensitive to both syntactic and semantic variables, the relative magnitudes of these effects vary a lot across these regions.
Furthermore, we found an asymmetry between the left and right hemispheres, with semantic and syntactic processing being more dissociated in the left hemisphere than in the right, and the left and right hemispheres showing respectively greater sensitivity to short and long contexts.
The use of information-restricted NLP models thus shed new light on the spatial organization of syntactic processing, semantic processing and compositionality. 
\end{abstract}

\section{Introduction}

Understanding the neural bases of language processing has been one of the main research efforts in the neuroimaging community for the past decades \citep[see, e.g.,][for reviews]{friedericibrain2011,binder2009review}. However, the complex nature of language makes it difficult to discern how the various processes underlying language processing are topographically and dynamically organized in the human brain, and therefore many questions remain open to this date. 

One central open question is whether semantic and syntactic information are encoded and processed jointly or separately in the human brain. Language comprehension requires to access word meanings (lexical semantics), but also to compose these meanings to construct the meaning of entire sentences. In languages such a English, semantic composition strongly depends on word order in the sentence -- for example, `The boy kissed the girl' has a different meaning compared to `The girl kissed the boy' although both sentences contain the exact same words. The brain constructs these different meanings conditionally on words order, which is the backbone of sentence processing, indicating how to combine the lexical meanings of its sub-parts. Importantly, meaning construction of new sentences would be roughly done in the same way if only the structure of the sentences remains the same (`The X kissed the Y'), independently of the lexical meanings of the single nouns in the sentences (`boy' and `girl'). This combinatorial property of language allows to construct meanings of sentences that we have never heard before and suggests that it might be computationally advantageous for the brain to have developed neural mechanisms for composition that are separate from those dedicated to the processing of lexico-semantic content. Such neural mechanisms for composition would be sensitive to only the abstract structure of sentences and would implement the syntactic rules according to which sentence parts should be composed.

Following related considerations, the dominant view over the past decades claimed that syntactic information is represented and processed in specialized brain regions, akin to the classic modular view \citep{fodormodularity1983,chomskymodular1984}. Neuronal modularity of language processing gained support from early lesion studies suggesting that syntactic processing takes place in localized and specialized brain regions such as Broca's area, showing double dissociations between syntactic and semantic processing \citep{goodglass1993understanding, caramazza1976dissociation}. Neuroimaging studies \citep{embick2000syntactic,vigliocco2000language,hashimoto2002specialization,garrard2004dissociation,friederici2006processing,palliercortical2011,hagoortnodes2014,GRODZINSKY2008474, shetreet2014processing, friederici2017language, Matchin2020} as well as simulation work on language acquisition and processing in artificial neural language models \citep{ullman2004contributions, o2006making, russin2019reilly, lakretz2019emergence, lakretz2021mechanisms} have provided further support to this view since then.

However, in parallel, an opposing view has argued that semantics and syntax are processed in a common distributed language processing system \citep{bates1989functionalism,dicklanguage2001,bates2002language}. Recent work in support of this view has raised concerns regarding the replicability of some of the early results from the modular view \citep{siegelman2019attempt} and provided evidence that semantic and syntactic processing in the language network might not be so easily dissociated from one another \citep{mollica2018high,fedorenkolack2020}. 

Neuroimaging studies, cited to defend one or the other view, have mainly relied on one of two methodological approaches: on the one hand, controlled experimental paradigms, which manipulate the words or sentences \citep{mazoyercortical1993,bottini1995righthemisphere,STROMSWOLD1996452,caplaneffects1998,palliercortical2011} and, on the other hand, naturalistic paradigms that make use of stimuli closer to what one could find in a daily environment. The former approach probes linguistic dimensions in one of the following ways: varying the presence or absence of syntactic or semantic information \citep{friedericidisentangling2009,friederici2003roleleftifg} or varying the syntactic structure difficulty or the semantic interpretation difficulty \citep[e.g.][]{cookeneural2001,friedericirole2009,kinnoneural2007,newmaneffect2010,SANTI20101285}.
However, the conclusions from such studies may be bounded to the peculiarity of the task and setup used in the experiment \citep{nastase2020}. To overcome these shortcomings, over the last years, researchers have become increasingly interested in data using ``Ecological Paradigms'', in which participants are engaged in more natural tasks, such as conversation or story listening \citep{regevselective2013,lernertopographic2011,wehbesimultaneously2014, nastasenarratives, pasquiouicml2022, lebel2022}. This avoids any task-induced bias and takes into consideration both lexical and supra-lexical levels of syntax and semantic processing. 
Integrating supra-lexical level information is essential for understanding language processing in the brain, because the lexical-semantic information of a word and the resulting semantic compositions depend on its context.

More recently, following advances in natural language processing, neural language models have been increasingly employed in the analysis of data collected from ecological paradigms. Neural language models are models based on neural networks, which are trained to capture joint probability distributions of words in sentences using next-word, or masked-word prediction tasks \cite[e.g.][]{elmandistributed1991,penningtonglove2014,devlinbert2019,radfordlanguage2019}. By doing so, the models have to learn semantic and syntactic relations among word tokens in the language. To study brain data collected from ecological paradigms, neural language models are presented with the same sentence stimuli, then, their activations (aka, embeddings) are extracted and used to fit and predict the brain data \citep{wehbesimultaneously2014,huthnatural2016,pasquiouicml2022,Caucheteux:2022}. This approach has led to several discoveries, such as wide networks associated with semantic processing uncovered by \cite{huthnatural2016} using word embeddings (see also \citealt{pereiratoward2018}), or context-sensitivity maps discovered by \cite{jainincorporating2018} and \cite{tonevainterpreting2019}.

Despite these advances and extensive neuroscientific and cognitive explorations, the neural bases of semantics, syntax and the integration of contextual information still remain debated. In particular, a central puzzle remains in the field: on the one hand, studies investigating syntax and semantics found vastly distributed networks when using naturalistic stimuli \citep{fedorenkolack2020,caucheteux:icml2021} and others found more localized activations for syntax, typically in inferior frontal and posterior temporal regions, when using constraint experimental paradigms e.g., \citep{palliercortical2011,matchin2017}. Thus, whether there is a hierarchy of brain regions integrating contextual information or the extent to which  syntactic information is independently processed from semantic information, in at least some brain regions, remains largely debated to date.

So far, insights from neural language models about this central puzzle were also rather limited. This is mostly due to the complexity of the models in terms of size, training and architecture. This complexity makes it difficult to identify how and what information is encoded in their latent representations, and how to use their embeddings to study brain function.

\citet{caucheteux:icml2021} used a neural language model, GPT-2, in an novel way to separate semantic and syntactic processing in the brain.
Specifically, using a pre-trained GPT-2 model, they built syntactic predictors by averaging the embeddings of words from sentences that shared syntactic but no semantic properties, and used them to identify syntactic-sensitive brain regions.
They defined as semantic-sensitive brain regions, the regions that were better predicted by the GPT-2's embeddings computed on the original text, compared to the syntactic predictors.
They observed that syntax and semantics, defined in this way, rely on a common set of distributed brain areas.

\cite{jainincorporating2018} used pre-trained LSTM models to study context integration. They varied the amount of context used to generate word embeddings, and obtained a map indicating brain regions' sensitivity to different sizes of context.

Here, we propose a new approach to tackle the questions of syntactic vs. semantic processing and contextual integration, by fitting brain activity with word embeddings derived from \textit{information-restricted} models.
By this, we mean that the models are trained on text corpora from which specific types of information (syntactic, semantic, or contextual) were removed.
We then assess the ability of these information-restricted models to fit brain activations, and compared it to the predictive performance of a neural model trained on the original dataset.

\begin{figure}
\begin{fullwidth}
  
  \includegraphics[width=0.95\linewidth]{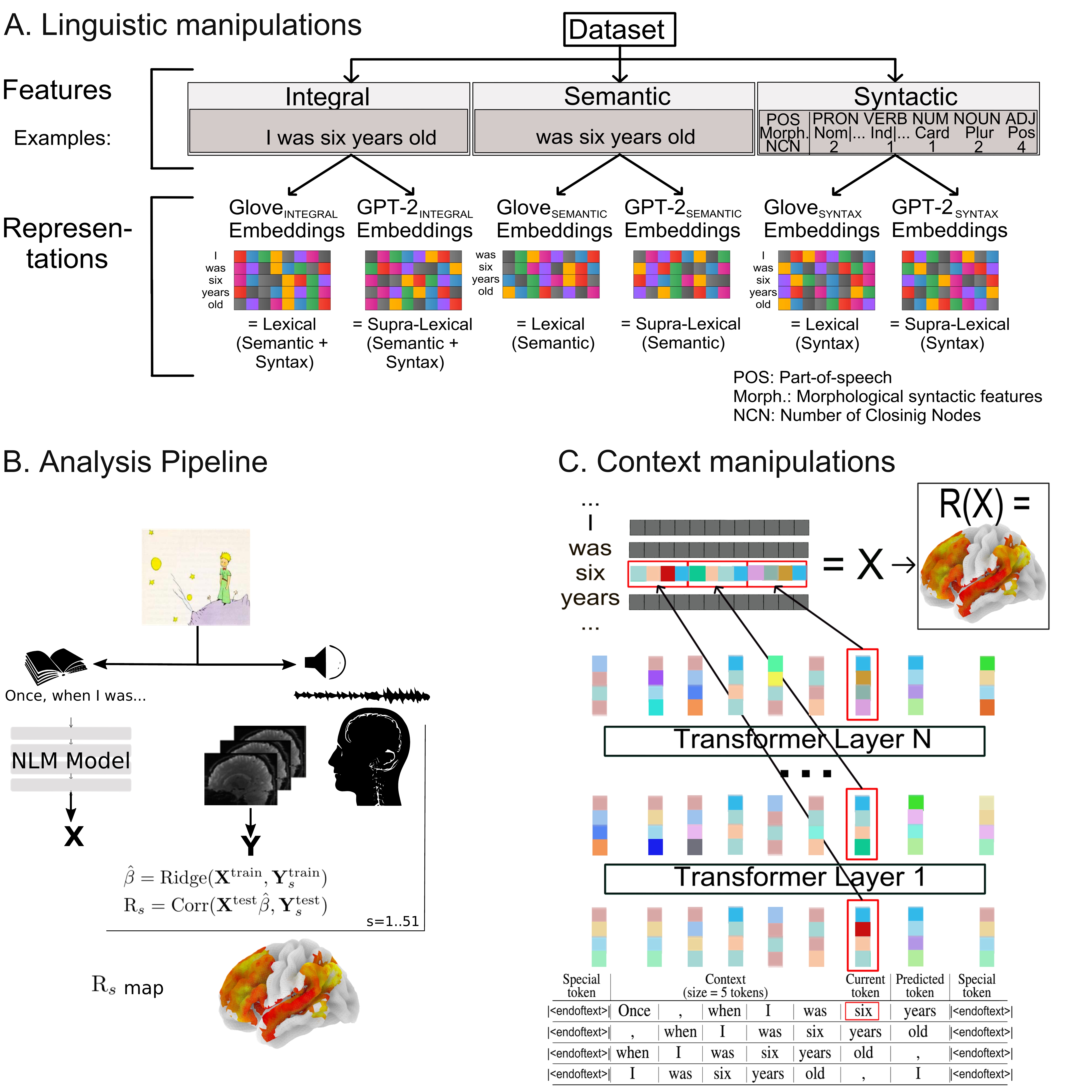}  \\

  \caption{\textbf{Experimental setup}
    \textbf{A)} A corpus of novels was used to create a dataset from which we extracted three different sets of features: (i) \textit{Integral features}, comprising all tokens (words+punctuation); (ii) \textit{Semantic features}, comprising only the content words; (iii) \textit{Syntactic features}, comprising syntactic characteristics (Part-of-speech, Morphological syntactic characteristics, Number of Closing Nodes) of all tokens. GloVe and GPT-2 models were trained on each feature space.
    \textbf{B)} fMRI scans of human participants listening to an audio-book were obtained. The associated text transcription was input to Neural models, yielding embeddings that were convolved with an haemodyamic kernel and fitted to brain activity using a Ridge-regression. Brain maps of cross-validated correlation between encoding models' predictions and fMRI time-series were computed. 
    \textbf{C)} To study sensitivity to context, a GPT-2 model was trained and tested on input sequences of bounded context length (5, 15 and 45). The resulting representations were then used to predict fMRI activity. }
  \label{Fig:Pipeline}
\end{fullwidth}
\end{figure}

More precisely, we created a text corpus of novels from the Gutenberg Project (\url{http://www.gutenberg.org}) and used it to define three different sets of features: (i) \textit{Integral features}, the full text from the corpus (ii) \textit{Semantic features}, the content words from the corpus; (iii) \textit{Syntactic features}, where each word and punctuation sign from the corpus is replaced by syntactic characteristics. We then trained two types of models on each feature space: a non-contextual model, Glove \citep{penningtonglove2014}, and a contextual model, GPT-2 \citep{radfordlanguage2019} (See Fig.~\ref{Fig:Pipeline}A).
The text transcription of the audio-book, to which participants listened in the scanner, was then presented to the neural language models from which we derived embedding vectors.
After fitting these embedded representations to fMRI brain data with linear encoding models, we computed the cross-validated correlations between the encoding models' predicted time courses and the observed time-series. In a first set of analyses, this allowed us to quantify the sensitivity to syntactic and semantic information in each voxel (Fig.~\ref{Fig:Pipeline}B).
In a second set of analyses, we identified brain regions integrating information beyond the lexical level. We first compared the contextual model (GPT-2) and the non contextual model (Glove), before investigating the brain regions processing short (5 words), medium (15 words) and long (45 words) contexts, using a non-contextualized GloVe model as a 0-context baseline (See Fig.~\ref{Fig:Pipeline}C.).

\section{Results}

\subsection*{Dissociation of syntactic and semantic information in embeddings}
\label{Decoding}
We first assessed the amount of syntactic and semantic information contained in the embedding vectors derived from GloVe and GPT-2 trained on the different sets of features. In order to do so, we trained logistic classifiers to decode either the semantic category or the syntactic category from the embeddings generated from the text of \emph{The Little Prince}.

\begin{figure}

  \includegraphics[width=\columnwidth]{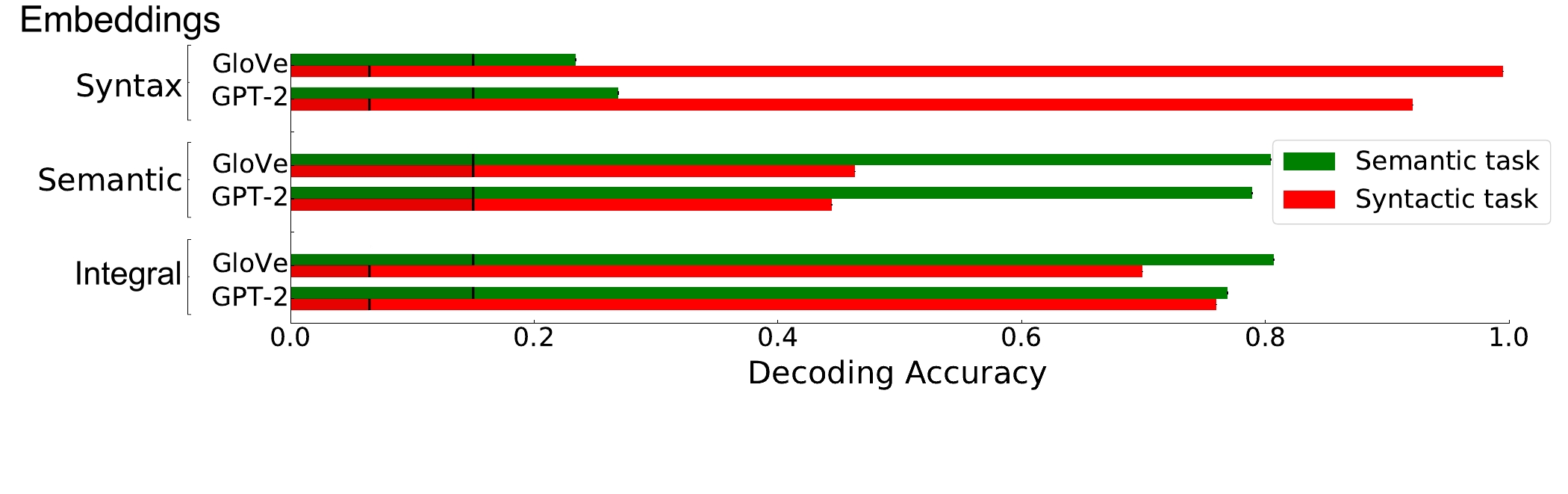}
  
  \caption{
    \textbf{Decoding syntactic and semantic information from words embeddings.} For each dataset and model type (Glove and GPT-2), logistic classifiers were set up to decode either the syntactic or the semantic categories of the words from the text of \emph{The Little Prince}. Chance-level was assessed using dummy classifiers and is indicated by black vertical lines.}

  \label{Figure:Decoding}

\end{figure}

The decoding performances of the logistic classifiers are displayed in Fig.\ref{Figure:Decoding}. The models trained directly on the integral features, that is, the intact texts, have relatively high performance on the two tasks (75\% in average for both GloVe and GPT-2). The models trained on the syntactic features performed well on the syntax decoding task (decoding accuracy >95\%), but are near chance-level on the semantic decoding task (decoding accuracy around 25\% with a chance-level at 16\%). Similarly, the models trained on the semantic features display good performance on the semantic decoding task (decoding accuracy greater than 80\%), but a relatively poorer decoding accuracy on the syntax decoding task (45\%, chance level: 16\%). These results validate the experimental manipulation by showing that syntactic embeddings essentially encode syntactic information and semantic embeddings essentially encode semantic information.

\subsection*{Correlations of fMRI data with syntactic and semantic embeddings}
\label{BrainFit}
  
\begin{figure}
  \includegraphics[width=\columnwidth]{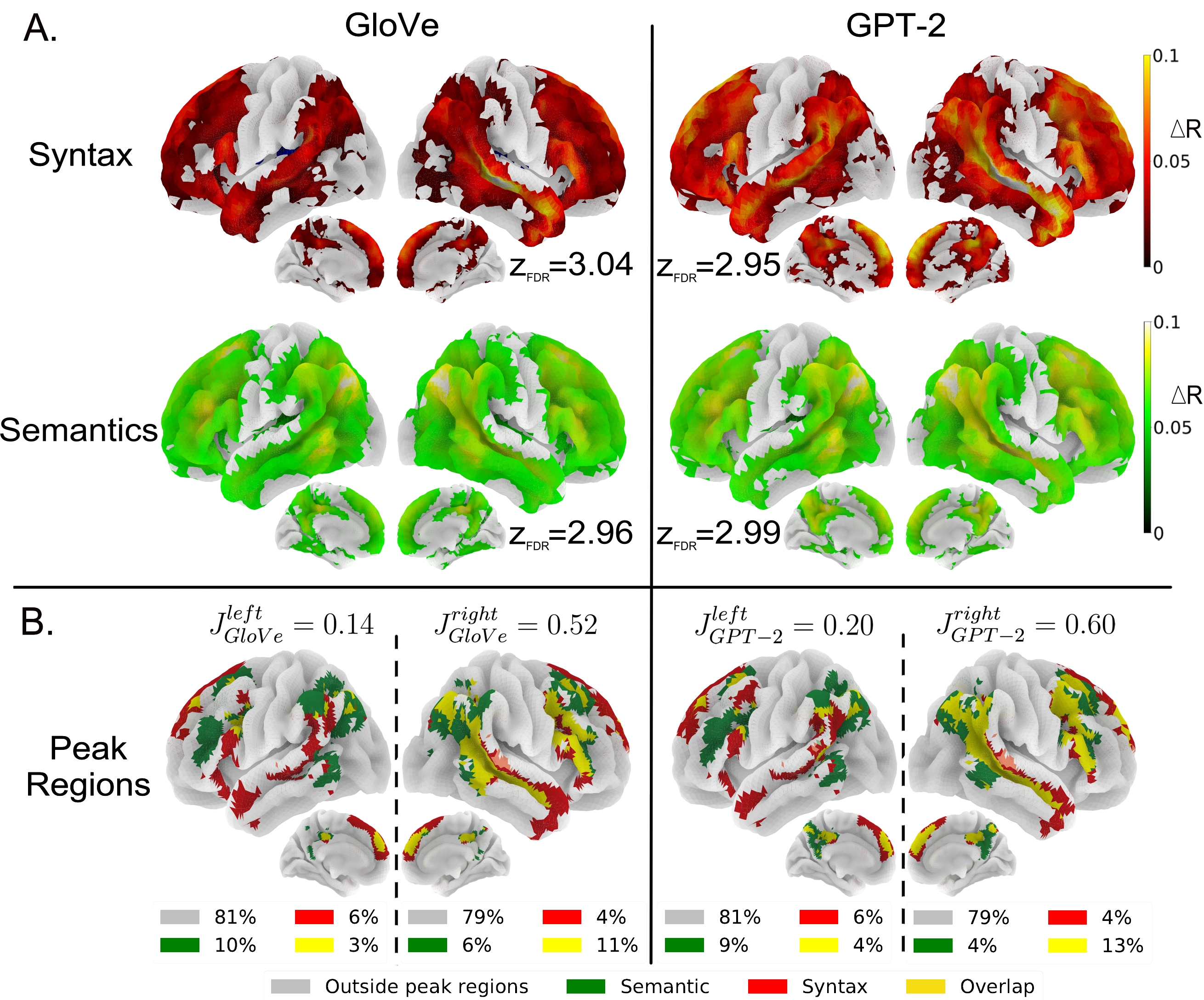} 
  \caption{
    \textbf{Comparison of the ability of GloVe and GPT-2 to fit brain
data when trained on either the semantic or the syntactic features.}
    \textbf{A)} Significant increase in R scores relative to the baseline model for GloVe (a non contextual model) and GPT-2 (a contextual model), trained either on the Syntactic features or on the Semantic features (voxel-wise thresholded group analyses; N=51 subjects; corrected for multiple comparisons with a FDR approach $p < 0.005$; for each figure $z_{FDR}$ indicates the significance threshold on the Z-scores).
    \textbf{B)} Bilateral spatial organisation of syntax and semantics highest R scores. Voxels whose R score belong in the 10\% highest R scores (in green for models trained on the semantic features, and in red for models trained on the syntactic features) are projected onto brain surface maps for GloVe and GPT-2 (overlap in yellow and other voxels in grey). Jaccard score for each hemisphere are computed, i.e. the ratio between the size of the intersection and the size of the union of semantics and syntax peak regions; the proportion of voxels of each category are displayed for each hemisphere and model.
  }
  \label{Figure:BrainFit}
\end{figure}

Our objective was to evaluate how well the embeddings computed from GloVe and GPT-2 on the syntactic and semantic features fit the fMRI signal in various parts of the brain.
For each model/features combination, we computed the increase in R score when the resulting embeddings were appended to a baseline model that comprised low-level variables (acoustic energy, word onsets and lexical frequency). This was done separately for each voxel. The resulting maps are displayed in Fig.\ref{Figure:BrainFit}A.

The maps reveal that semantic and syntactic feature-derived embeddings from GloVe or GPT-2 significantly explain the signal in a set of bilateral brain regions including frontal and temporal regions, as well as the Temporo-parietal junction, the Precuneus and Dorso-Medial Prefrontal Cortex (dMPC). The classical left-lateralized language network, which includes the Inferior Frontal Gyrus (IFG) and the Superior Temporal Sulcus (STS), is entirely covered. Overall, a vast network of regions is modulated by both semantic and syntactic information.

Nevertheless, detailed inspection of the maps shows different R score distribution profiles (see Appendix 1-\nameref{Appendix:RScoreDistribution} Appendix 1-Fig.\ref{Appendix:Figure:RScoreDistribution}).
For example, syntactic embeddings yield the highest fits in the Superior Temporal Lobe, extending from the Temporal Pole (TP) to the Temporo-Parietal Junction (TPJ), as well as the Inferior Frontal Gyrus (IFG, BA-44 and 47), the Superior Frontal Gyrus (SFG), the Dorso-Medial Prefrontal Cortex (dMPC) and the posterior Cingulate cortext (pCC). Semantic embeddings, on the other hand, show peaks in the posterior Middle Temporal Gyrus (pMTG), the Angular Gyrus (AG), the Inferior Frontal Sulcus (IFS), the dMPC and the Precuneus/pCC.

\subsection*{Regions best fitted by semantic or syntactic embeddings}
\label{PeakRegions}

As noticed above, despite the fact that the regions fitted by semantic and syntactic embeddings essentially overlap (Fig.\ref{Figure:BrainFit}A), the areas where each model has the highest R scores differ.
To better visualize the maxima from these maps, we selected, for each of them, the 10\% of voxels having the highest R scores. Thresholding at the 90-th percentile of the distributions (threshold values displayed in Appendix 1-Fig.\ref{Appendix:Figure:RScoreDistribution}) produces the maps presented in Fig.\ref{Figure:BrainFit}B. 

A first observation is that the number of supra-threshold voxels is quite similar in the left (19\%) and right (21\%) hemispheres, whether GPT-2 or Glove is considered, showing that during the processing of natural speech, both syntactic and semantic features modulate activations in both hemispheres to a similar extent. 
The regions involved include, bilaterally, the TP, the STS, the IFG and IFS, the DMPC, the pMTG, the TPJ, the Precuneus and pCC.

One noticeable difference between the two hemispheres, apparent in Fig.\ref{Figure:BrainFit}B, concerns the \emph{overlap} between the semantic and syntactic peak regions: it is stronger in the right than in the left hemisphere. To assess this overlap, we computed the Jaccard indices (see \nameref{Method:Jaccard}) between voxels modulated by syntax and voxels modulated by semantics. The Jaccard indices were much larger in the right hemisphere ($J^{right}_{GloVe}=0.52$ and $J^{right}_{GPT-2}=0.60$) than in the left ($J^{left}_{GloVe}=0.14$ and $J^{left}_{GPT-2}=0.20$).

The left hemisphere displayed distinct peak regions for semantics and syntax; syntax involving the STS, the pSTG, the anterior TP, the IFG (BA-44/45/47) and the MFG, while semantics involves the pMTG, AG, the TPJ and the IFS. We only observe overlap in the upper IFG (BA-44), AG and posterior STS. On medial faces, semantics and syntax share peak regions in the Precuneus, the pCC and the DMPC. In the right hemisphere, syntax and semantics share the STS, pMTG and most frontal regions, with only syntax-specific peak regions in the TP and SFG and semantics-specific peak regions in the TPJ.

Overall, this shows that the neural correlates of syntactic and semantic features appear more separable in the left than in the right hemisphere .

\begin{figure}
  \includegraphics[width=\columnwidth]{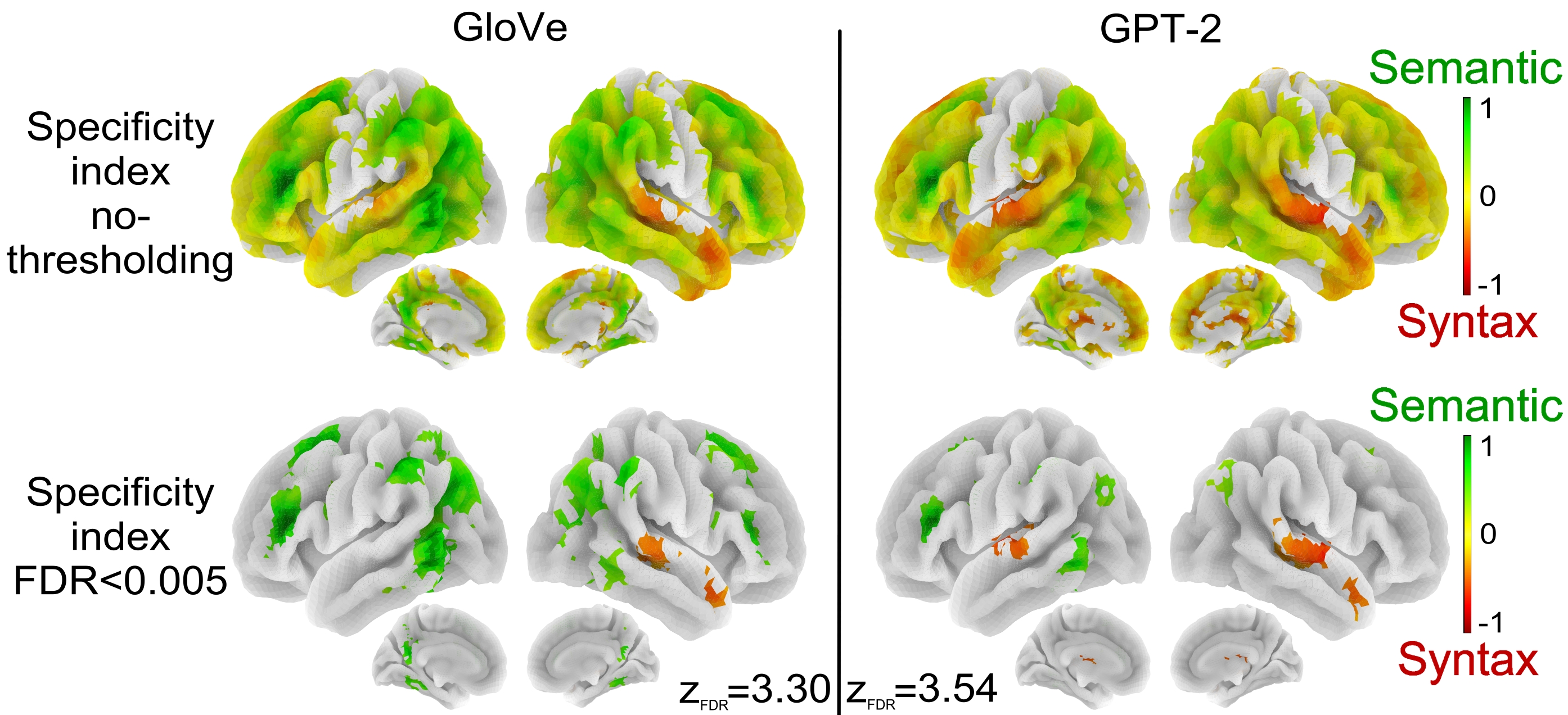} 
  \caption{
    \textbf{Voxels' sensitivity to syntactic and semantic embeddings.}
    Voxels' specificity indexes are projected onto brain surface maps reflecting how much semantic information helps to better fit the time-courses of a voxel compared to syntactic information; the greener the more the voxel is categorized as a semantic voxel, the redder the more the voxel is categorized as a syntactic voxel. Yellow regions are brain areas where semantic and syntactic information lead to similar R score increases.
    The top row displays specificity indexes in voxels where there was a significant effect for semantic or syntactic embeddings in Fig.\ref{Figure:BrainFit}A.
    The bottom row is the voxel-wise thresholded group analyses; N=51 subjects; corrected for multiple comparisons with $FDR<0.005$ (for each figure $z_{FDR}$ indicates the significance threshold on the Z-scores).
    }
  \label{Figure:DiffRatio}
\end{figure}

\subsection*{Gradient of sensitivity to syntax or semantics}
\label{DiffRatio}

The analyses presented above revealed a large distributed network of brain regions sensitive to both syntax and semantics but with varying local sensitivity to both conditions.

We further investigated these differences by defining a \textit{specificity index} that reflects, for each voxel, the logarithm of the ratio between the R scores derived from the semantic and the syntactic embeddings (see \nameref{Method:SpecificityIndex}).
A score of $x$ indicates that the voxel is $10^{x}$-times more sensitive to semantics compare to syntax if $x>0$ (green), and conversely, the voxel is $10^{-x}$-times more sensitive to syntax compare to semantics if $x<0$ (red).
Voxels with specificity indexes close to 0, are colored in yellow and show equal sensitivity to both conditions.
Specificity indexes are plotted on surface maps in Fig.\ref{Figure:DiffRatio}.
The top row shows the specificity index of voxels where there was a significant effect for syntactic or for semantic embeddings in Fig.\ref{Figure:BrainFit}A, while the bottom row shows group specificity indexes corrected for multiple comparison using an FDR-correction of 0.005 (N=51).

The top row of Fig.\ref{Figure:DiffRatio} shows that voxels that are more sensitive to Syntax include, bilaterally, the anterior Temporal Lobes (aTL), the STG, the Supplementary Motor Area (SMA), the MFG and sub-parts of the IFG.
Voxels more sensitive to Semantics are located in the pMTG, the TPJ/AG, the IFS, SFS and the Precuneus.
Voxels sensitive to both types of features are located in the posterior STG, the STS, the dMPC, the CC, the MFG and in the IFG.

More specifically, in Fig.\ref{Figure:DiffRatio} bottom, one can observe significantly low ratios (in favor of the syntactic embeddings) in the STG, aTL and pre-SMA, and significantly large ratios (in favor of the semantic embeddings) in the pMTG, the AG and the IFS.
Specificity index maps are consistent with the maps of R score differences between semantic and syntactic embeddings for Glove and GPT-2 (see Appendix 1-Fig.\ref{Appendix:Figure:DiffRatio}), but provide more insights into the relative sensitivity to syntax and semantics. 
These maps highlight that some brain regions show stronger responses to the semantic or to the syntactic condition even when they show sensitivity to both.

\subsection*{Unique contributions of syntax and semantics}
\label{UniqueCorrelation}

The previous analyses allowed us to quantify the amounts of brain signal explained by the information encoded in various embeddings.
Yet, when two embeddings explain the same amount of signal, that is, have similar R score, it remains to be clarified whether they hinge on information represented redundantly in the embeddings or information specific to each embedding.
To address this issue, we analyzed the additional information brought by each embedding on top of the other one. To this end, we evaluated correlations that are uniquely explained by the semantic embeddings compared to the syntactic embeddings, and conversely.

To quantify the unique contribution of each feature space to the prediction of the fMRI signal, we first estimated the Pearson correlation explained by the embeddings learned from the individual feature space - e.g., using only syntactic embeddings or semantic embeddings. We then assessed the correlation explained by the concatenation of embeddings derived from different feature spaces - e.g., concatenating syntactic and semantic embedding vectors \citep{deheerhierarchical2017}.

Because it can identify single voxels whose responses can be partly explained by different feature spaces, this approach provides more information than simple subtractive analyses that estimate the R score difference per voxel (see Appendix 1-Fig.\ref{Appendix:Figure:DiffRatio}).

\begin{figure}
  \includegraphics[width=\columnwidth]{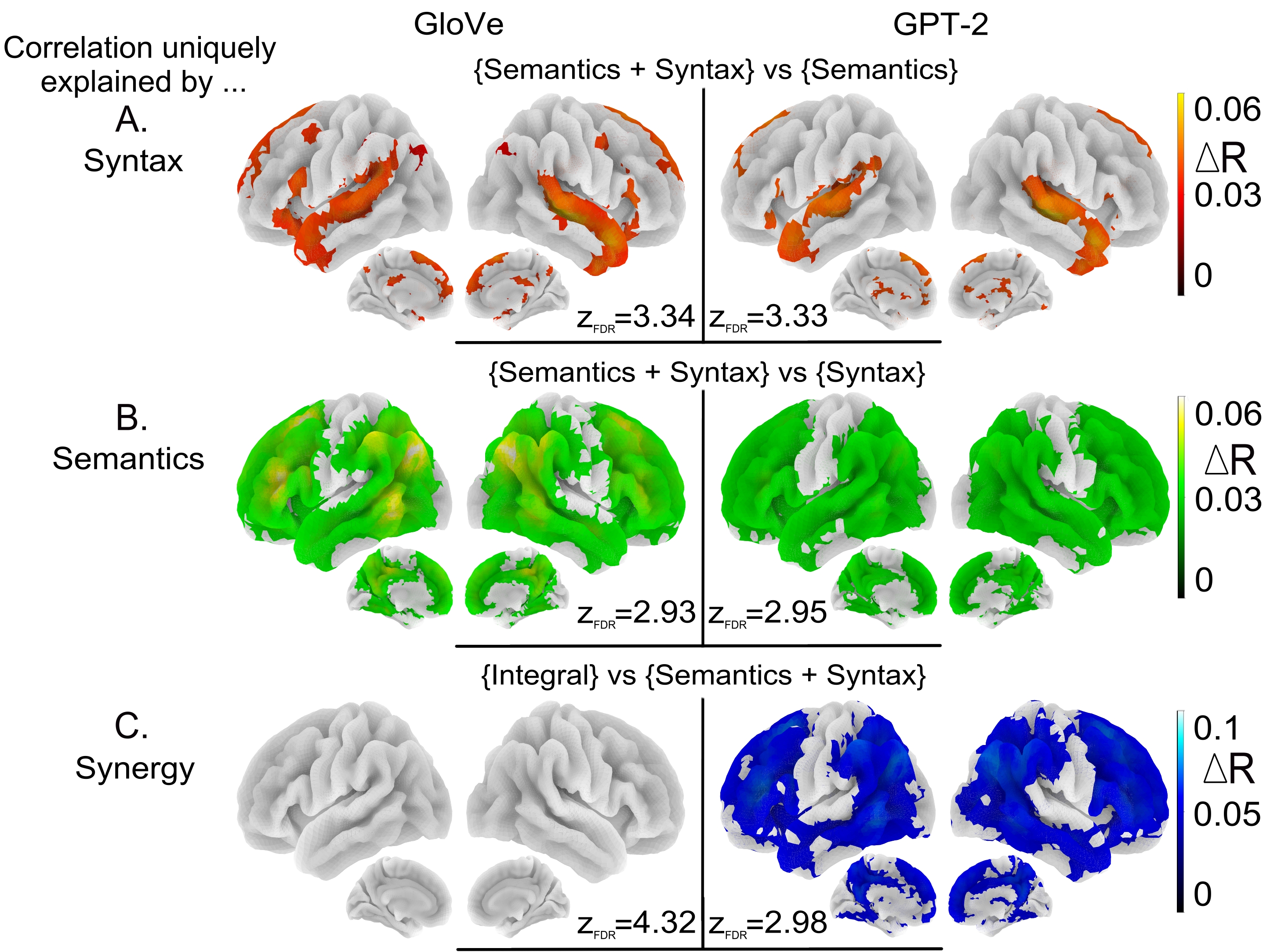}
  \caption{
    \textbf{Correlation uniquely explained by each embeddings}
    \textbf{A)} Increase in R scores relative to the semantic embeddings when concatenating semantic and syntactic embeddings in the encoding model.
    \textbf{B)} Increase in R scores relative to the syntactic embeddings when concatenating semantic and syntactic embeddings in the encoding model.
    \textbf{C)} Increase in R scores relative to the concatenated semantic and syntactic embeddings for the integral embeddings.
    These maps are voxel-wise thresholded group analyses; N=51 subjects; corrected for multiple comparisons with a FDR approach $p < 0.005$; for each figure $z_{FDR}$ indicates the significance threshold on the Z-scores.
}
  \label{Figure:UniqueCorrelation}
\end{figure}

Syntactic embeddings (Fig.\ref{Figure:UniqueCorrelation}A) uniquely explained brain data in localized brain regions: the STG, the TP, the pre-SMA and in the IFG, with R scores increases of about 5\%.

Semantic embeddings (Fig.\ref{Figure:UniqueCorrelation}B) uniquely explained signal bilaterally in the same wide network of brain regions as the one highlighted in Fig.\ref{Figure:BrainFit}A, including frontal and temporo-parietal regions bilaterally as well as the Precuneus and pCC medially, with similar R scores increases around 5\%.

This suggests that even if most of the brain is sensitive to both syntactic and semantic conditions, syntax is preferentially processed in more localized regions than semantics which is widely distributed.

\subsection*{Synergy between syntax and semantics}
\label{Synergy}

To probe regions where the joint effect of syntax and semantics is greater than the sum of the contributions of these features, we compared the R scores of the embeddings derived from the integral features with the R scores of the encoding models concatenating the semantic and syntactic embeddings (see Fig.\ref{Figure:UniqueCorrelation}C).

For the embeddings obtained with GloVe, this analysis did not reveal any significant effect. For the embeddings obtained with GPT-2, significant effects were observed in most of the brain, but with higher effects in the semantic peak regions: pMTG, TPJ, AG and in frontal regions.

\subsection*{Integration of contextual information}
\label{Context}

\begin{figure}[!ht]
    \resizebox{\textwidth}{!}{%
    \includegraphics{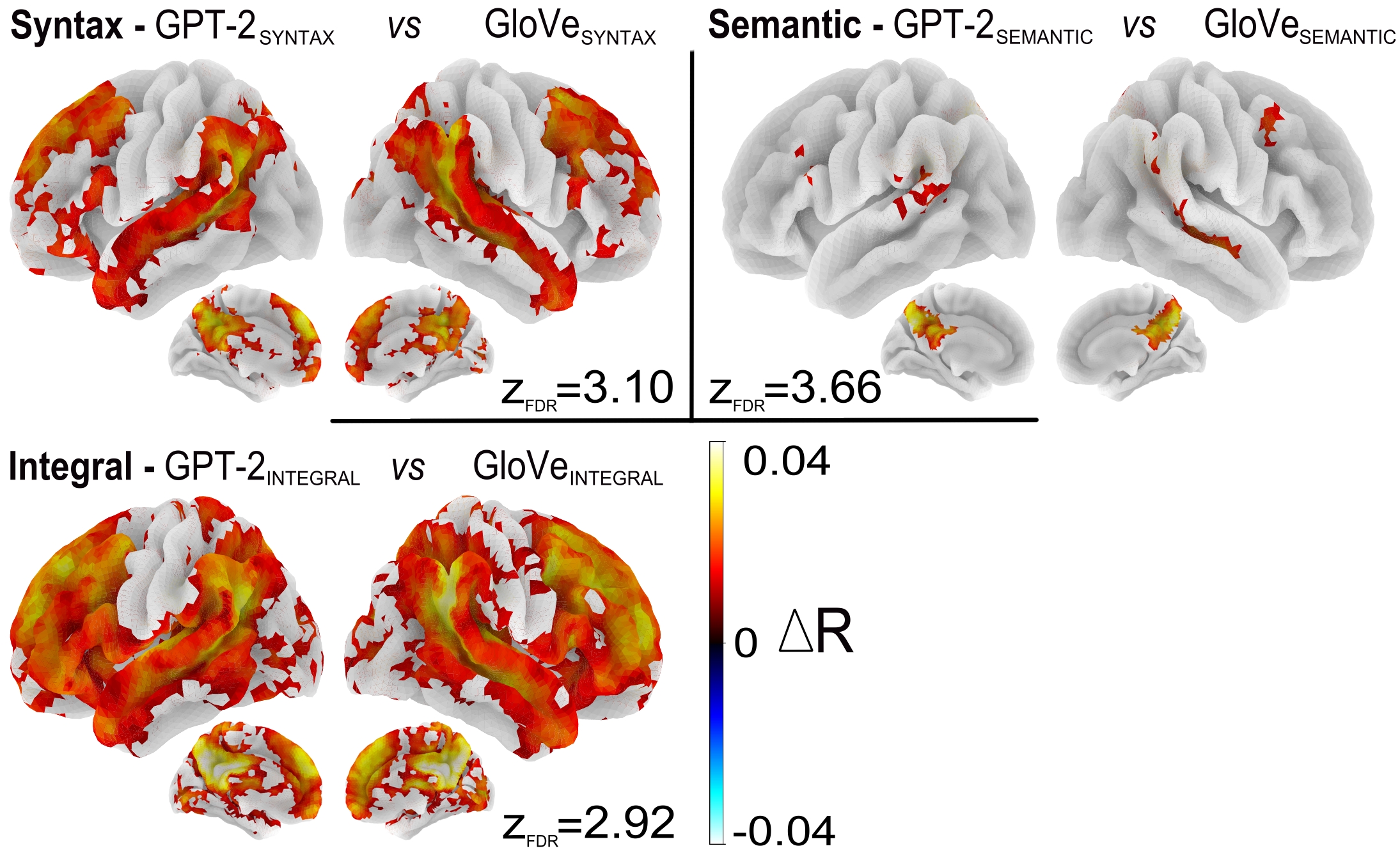}}
    \caption{\textbf{Comparison of lexical and supra-lexical processing levels.}
        Brain regions that are significantly better predicted by GPT-2 (in red) compared to Glove, when trained on syntactic features (top left), semantic features (top right) and integral features (bottom left).
        Maps are voxel-wise thresholded group analyses; N=51 subjects; corrected for multiple comparisons with a FDR approach $p < 0.005$; for each figure $z_{FDR}$ indicates the significance threshold on the Z-scores.
    }
    \label{Figure:SupraLexical}
\end{figure}

To further examine the effect of context, we compared GPT-2, the supra-lexical model which takes context into account, to GloVe, a purely lexical model. The differences in R scores between the two models, trained on each of the three datasets are presented in Fig.\ref{Figure:SupraLexical}.

GPT-2 embeddings elicit stronger R scores than GloVe.
The difference spreads over wider regions when the models were trained on syntax compared to semantics (see Fig.\ref{Figure:SupraLexical} top left and right).
The comparison for syntax led to significant differences bilaterally in the STS/STG, from the Temporal Pole to the TPJ, in superior, middle and inferior frontal regions, and medially in the pCC and dMPC.
For semantics, the comparison only led to significant differences in the Precuneus, the right STS and posterior STG.
Fig.\ref{Figure:SupraLexical} (bottom left) shows the comparison between GPT-2 and GloVe when trained on the Integral features.
Given that both semantic and syntactic contextual information were available to GPT-2, these maps reflect the regions that benefit from context during story listening.

To show that context has an effect is one thing, but different brain regions are likely to have different integration window's size.
To address this question, we developed a fixed-context window training protocol to control for the amount of contextual information used by GPT-2 (Fig.\ref{Fig:Pipeline}C). 
We trained models with short (5 tokens), medium (15 tokens) and long (45 tokens) range windows sizes.
This ensures that GPT-2 was not sampling out of the learnt distribution at inference, and not using more context than what was available in the context window.

Comparing GPT-2 with 5 tokens to GloVe (0-size context) highlighted a large network of frontal and temporo-parietal regions. Medially, it included the Precuneus, the pCC and the DMPC (Fig.\ref{Figure:SupraLexical}, short).
Short context-sensitivity showed peak effects in the Supramarginal gyri, the pMTG and medially in the Precuneus and pCC.
Counting the number of voxels showing significant short-context effects highlighted an asymmetry between the left and right hemisphere with 1.6 times more significant voxels in the left hemisphere compared to the right.
Contrasting a GPT-2 model using 15 tokens of context (the average size of a sentence in \emph{The Little Prince}) versus a GPT-2 model using only 5 tokens, yielded localized significant differences in the SFG/SFS, the TP, MFG and STG near Heschl's gyri and medially in the Precuneus and pCC (Fig.\ref{Figure:SupraLexical}, Medium).
The biggest medium context effects included the left MFG, the right SFG and DMPC and bilaterally the Precuneus and pCC.
Finally, contrasting models using respectively 45 and 15 tokens of context revealed 2.8 times as many significant differences in the right hemisphere as in the left.
Significant effects were the highest bilaterally and medially in the pCC, followed, in the right hemisphere, by the Precuneus, the DMPC, MFG, SFG, STS and TP (see Fig.\ref{Figure:SupraLexical}, bottom).

\begin{figure}[!ht]
    \resizebox{\textwidth}{!}{%
    \includegraphics{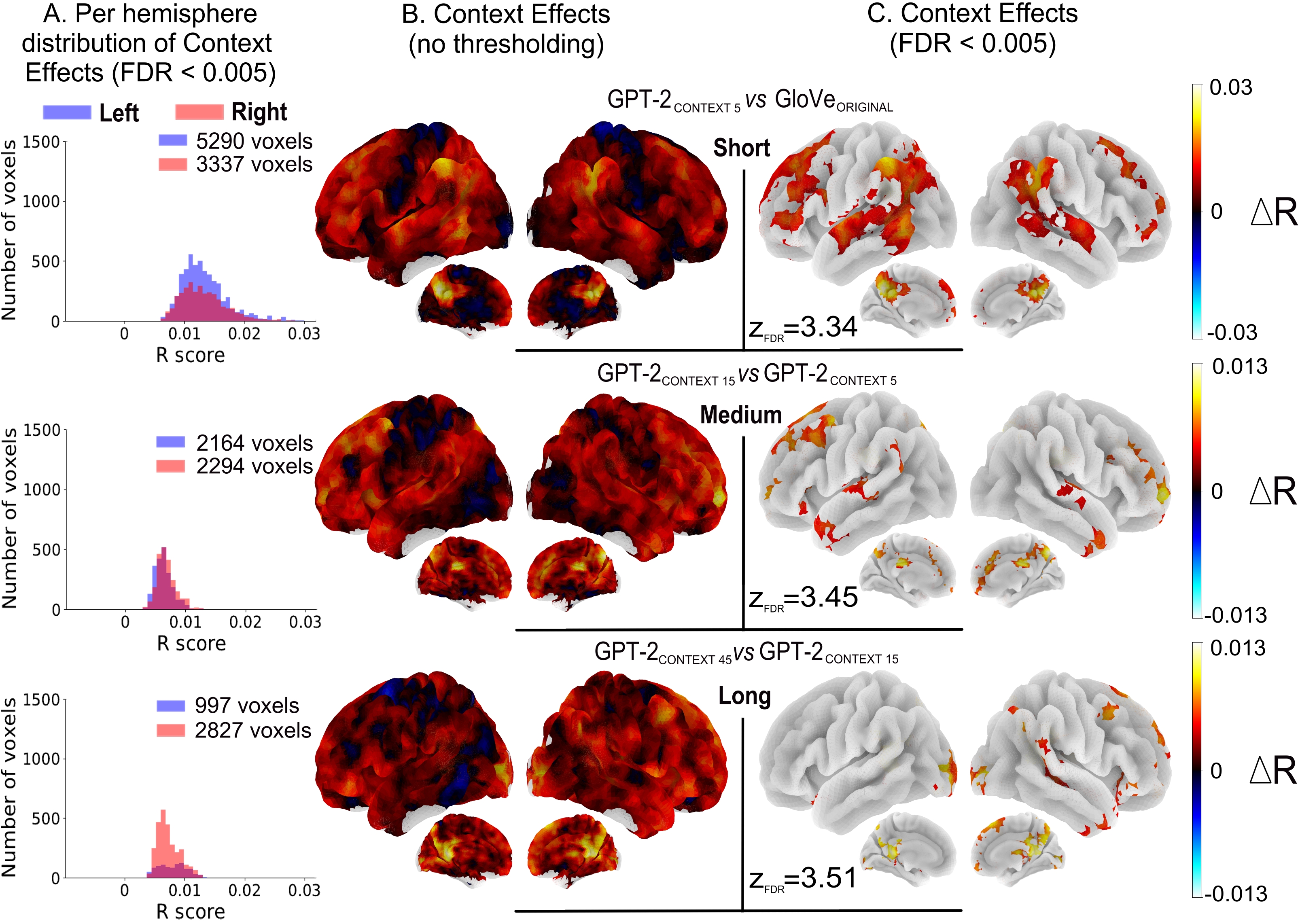}}
    \caption{\textbf{Integration of context at different levels of language processing.}
        \textbf{A)} Per hemisphere histograms of significant context effects after group analyses (N=51 subjects); thresholded at p<0.005 voxel-wise, corrected for multiple comparisons with the FDR approach.
        \textbf{B)} Uncorrected group averaged surface brain maps representing R scores increases when fitting brain data with models leveraging increasing sizes of contextual information.
        \textbf{C)} Corrected group averaged surface brain maps representing R scores increases when fitting brain data with models leveraging increasing sizes of contextual information; thresholded at p<0.005 voxel-wise, corrected for multiple comparisons with the FDR approach (for each figure $z_{FDR}$ indicates the significance threshold on the Z-scores).
        \textbf{(top row)} Comparison of the model trained with 5 tokens of context (GPT-$2_{Context-5}$) with the non-contextualized GloVe. 
        \textbf{(middle row)} Comparison of the models respectively trained with 15 (GPT-$2_{Context-15}$) and 5 (GPT-$2_{Context-5}$) tokens of context.
        \textbf{(bottom row)} Comparison of the models respectively trained with 45 (GPT-$2_{Context-45}$) and 15 (GPT-$2_{Context-15}$) tokens of context.
    }
    \label{Figure:Context}
\end{figure}

Taken together, our results show 1) that syntax dominantly determines the integration of contextual information, 2) that a bilateral network of frontal and temporo-parietal regions is modulated by short context, 3) that short-range context integration is preferentially located in the left hemisphere, 4) that the right hemisphere is involved in the processing of longer context sizes, and finally 5) that medial regions (Precuneus and pCC) are core regions of context integration, showing context effects at all scales.

\section{Discussion}

\begin{figure}[!ht]
\includegraphics[width=\columnwidth]{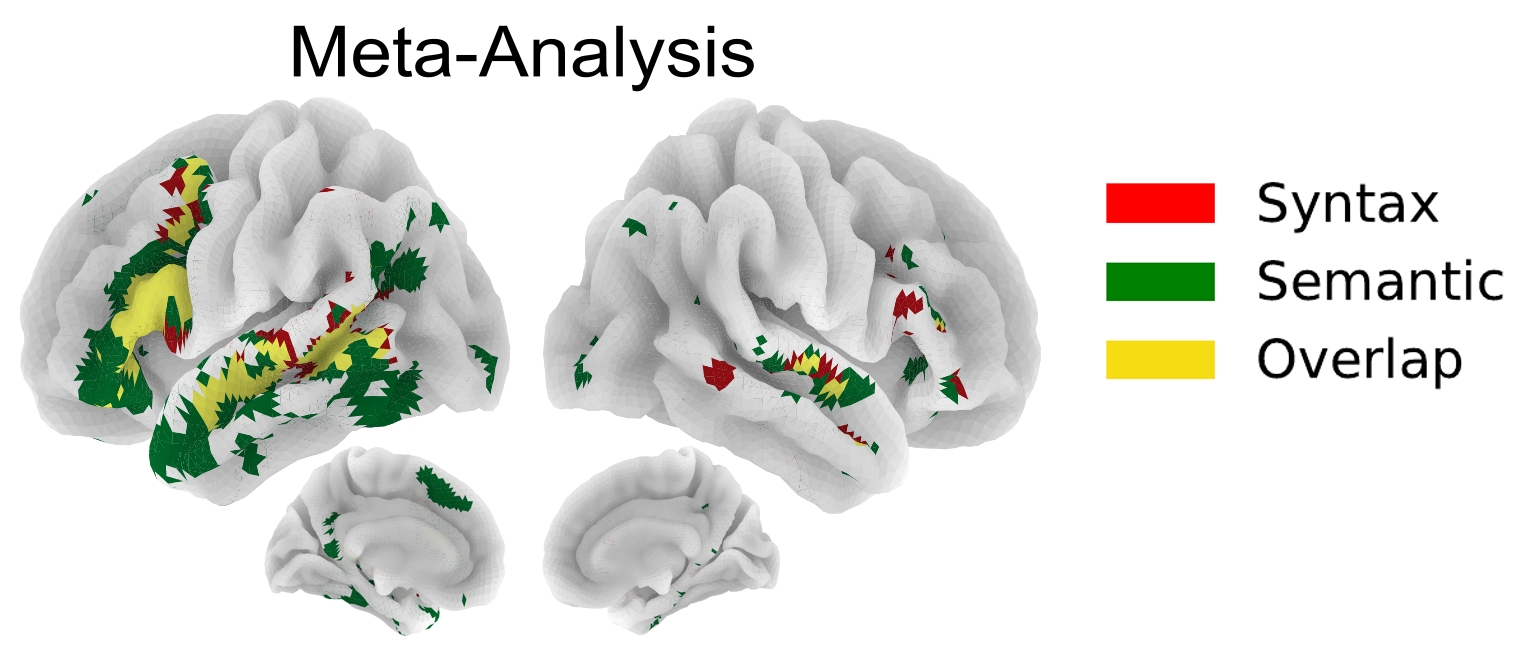} 
\caption{\textbf{Association maps for the terms ``semantic'' and ``syntactic'' in a meta-analysis using Neurosynth} (\protect\url{http://neurosynth.org}) The association test map for syntactic (resp. semantic) displays voxels that are reported more often in articles that include the term syntactic (resp. semantic) in their abstracts than articles that do not (FDR correction of 0.01).}

\label{Figure:MetaAnalysis}
\end{figure}

Language comprehension in humans is a complex process, which involves several interacting sub-components (word recognition, processing of syntactic and semantic information to construct sentence meaning, pragmatic and discourse inference, ...) \citep[e.g.]{Jackendoff2002JACFOL2}.
Discovering how the brain implements these processes is one of the major goals of neurolinguistics.
A lot of attention has been devoted, in particular, to the syntactic and semantic components \citep[][for reviews]{friedericineurobiology2017,binder_neurobiology_2011} and the extent to which they are implemented in (practically) distinct or identical regions is still debated \citep[e.g.][]{fedorenkolack2020}.
In Fig.\ref{Figure:MetaAnalysis}, we present the outcome of a meta analysis of the literature based on the search for the keywords 'syntactic' and 'semantic' in the Neurosynth database (see \nameref{Methods:MetaAnalysis}).
This analysis, albeit somewhat simplistic, reveals the brain regions most often associated with syntax and semantics.

It must be noted that a fair proportion of the studies included in the meta analysis relied on controlled experimental paradigms with single words or sentences, based on the manipulation of complexity or violations of expectations.
To study language processing in a more natural way, several recent studies have presented naturalistic texts to participants, and have analyzed their brain activations using Artificial Neural Language Models \citep[e.g.][]{pereiratoward2018,huthnatural2016,schrimpfartificial2020,pasquiouicml2022}.
These models are known to encode some aspects of semantics and syntax \citep[e.g.][]{penningtonglove2014,hewitt2019,lakretz2019emergence}.
In the current work, to further dissect brain activations into separate linguistic processes,  we trained NLP models on a corpus from which we selectively removed syntactic, semantic or contextual information and examined how well these information-restricted models could explain fMRI signal recorded from participants who had listened to an audiobook.
The rationale was to highlight brain regions representing syntactic and semantic information, at the lexical and supralexical levels (comparing a lexical model GloVe, and a contextual one, GPT-2).
Additionaly, by varying the amount of context provided to the supralexical model, we sought to identify the brain regions sensitive to different context sizes (see \cite{jainincorporating2018} for a similar analysis).

Whether models were trained on syntactic features or on semantic features, they fit fMRI activations in a wide bilateral network which goes beyond the classic language network comprising the IFG and temporal regions: it also includes most of the dorso lateral and medial prefrontal cortex, the inferior parietal cortex, and on the internal face, the precuneus and posterior cingulate cortex (see Fig.\ref{Figure:BrainFit}). 
Nevertheless, the regions \emph{best} predicted by syntactic features on the one hand, and semantic features on the other hand, are not exactly the same.
While they overlap quite a lot in the right hemisphere, they are more dissociated in the left hemisphere Fig.\ref{Figure:BrainFit}, panel B).
In addition, the relative sensitivity to syntax and semantics varies from region to region, with syntax predominating in the temporal lobe (see Fig.\ref{Figure:DiffRatio}).
Elimination of shared variance between syntactic and semantic features confirmed that pure syntactic effects are restricted to STG/STS, bilaterally, IFG, and pre-SMA, while pure semantic effects occur throughout the network (Fig.\ref{Figure:UniqueCorrelation} A-B).

The comparison between the supralexical model (GPT-2) and the lexical one (GloVe), revealed brain regions involved in compositionality  (Fig.\ref{Figure:SupraLexical}) and a synergy between syntax and semantics that arises only at the supralexical level (Fig.\ref{Figure:UniqueCorrelation}C). 
Finally, analyses of the influence of the size of context provided to GPT-2 when computing word embeddings, show that (1) a bilateral network of fronto-temporo-parietal regions is sensitive to short context, that (2) there is a dissociation between the left and right hemispheres, respectively associated with short-range and long-range context integration, and finally that (3) the medial Precuneus and posterior Cingulate gyri show the highest effects at every scale, hinting at an important role in large context integration (Fig.\ref{Figure:Context}).

\subsection{Models trained on semantic and syntactic features fit brain activity in a widely distributed network, but with varying relative degrees.}

When trained on the integral corpus, that is on the integral features, both the lexical (GloVe) and contextual (GPT-2) models captured brain activity in a large \emph{extended language network} (Appendix 1-Fig.\ref{Appendix:Figure:BrainFit}).
This large extended language network goes beyond the \emph{core} language network, that is, the left IFG and temporal regions, encompassing homologous areas in the right hemisphere, the dorsal prefrontal regions, both on the lateral and medial surfaces, as well as in the inferior parietal, Precuneus and posterior Cingulate.
The result is consistent with the ones from previous studies that have looked at brain responses to naturalistic text, whether analysed with NLP models \citep[e.g.][]{huthnatural2016,pereira_toward_2018,jainincorporating2018,caucheteux:icml2021} or not \citep{lernertopographic2011,chang_information_2022}.

The Precuneus/pCC, inferior parietal and dorsomedial prefrontal cortex are part of the Default Mode Network (DMN) 
\citep{raichle_brains_2015}. 
The same areas are actually also relevant in language and high-level cognition. 
For example, early studies examining the role of coherence during text comprehension had pointed out the same regions \citep{ferstl_role_2001,xu_language_2005}: coherent discourses elicit stronger activations than incoherent ones.
Recent work by \citep{chang_information_2022} has revealed that the DMN is the last stage in a temporal hierarchy of processing naturalistic text, integrating information on the scale of paragraphs and narrative events, see also \citep{simony2016,BALDASSANO2017709}. 
These regions are not language-specific though, as they have been shown to be activated during various theory of mind tasks, relying on language or not, and have thus also been dubbed the ``Mentalizing network'' \citep{mar_neural_2011,baetens_involvement_2014}. 

Models trained on the information-restricted semantic and syntactic features fit signal in this widely distributed network (Fig.\ref{Figure:BrainFit}A). This is in agreement with \cite{caucheteux:icml2021} and \cite{fedorenkolack2020} who, using very different approaches, found that syntactic predictors modulated activity throughout the language network. 
\cite{caucheteux:icml2021} first constructed new texts that matched, as well as possible, the text presented to participants in terms of their syntactic properties. The lexical items being different, the semantics of the new texts bear little relation with the original text. Then, using a pre-trained version of GPT-2, the authors obtained embeddings from these new texts and averaged them to create syntactic predictors. 
They found that these syntactic embeddings fitted a network of regions (ibid. Fig5D) similar to the one we observed (Fig.\ref{Figure:BrainFit}A).
Further, defining the effect of semantics as the difference between the scores obtained from the embeddings from the original text, and the scores from the syntactic embeddings, \cite{caucheteux:icml2021} observed that semantics had a significant effect throughout the same network (ibid. Fig5G).

Should one conclude that syntax and semantics equally modulate the entire language network?
Our results reveal a more complex picture. 
Figure \ref{Figure:DiffRatio} presents a semantics vs syntax specificity index map, showing higher sensitivity to syntax in the STG and anterior temporal lobe, whereas the parietal regions are more sensitive to semantics, consistent with \cite{binder2009review}.
Another point to take in consideration is that syntactic and semantic features are not perfectly orthogonal. 
Indeed, the logistic decoder trained on the embeddings from the semantic dataset was better than chance at recovering syntactic features (Fig.\ref{Figure:Decoding}), and vice versa. 
This might be due, for example, to the fact that some features like gender or number are present in both datasets, explicitly in the syntactic dataset and implicitly in the semantic dataset. 
To focus on the unique contributions of syntax and semantic, we remove the shared variance from the syntactic and semantic models using model comparisons (Fig.\ref{Figure:UniqueCorrelation}).

\subsection*{``Pure'' semantic but not ``pure'' syntactic features modulate activity in a wide set of brain regions.}

The unique effect of semantics, when its shared component with syntax was removed, remains widespread (Fig.\ref{Figure:UniqueCorrelation}B).
This is consistent with the notion that semantic information is widely distributed over the cortex, an idea  popularized by  embodiment theories \citep{Hauk2004,pulvermuller2013}, but which was already supported by the neuropsychological observations revealing domain-specific semantic deficits in patients \citep{damasio2004}.

On the other hand the ``pure'' effect of syntax ``shrinked'' to the STG and aTL (bilaterally), the IFG (on the left) and the pre-SMA (Fig.\ref{Figure:UniqueCorrelation}A).
The left IFG and STG/STS have previously been implicated in syntactic processing \cite[e.g.]{friedericineurobiology2017,friedericibrain2011}, and this is confirmed by the new approach employed here.
Note that we are not claiming that these regions are specialized for syntactic processing only.
Indeed they also appear to be sensitive to the ``pure'' semantic component  (Fig.\ref{Figure:UniqueCorrelation}B).

\subsection*{The contributions of the right hemisphere.}

A striking feature of our results is the strong involvement of the right hemisphere.
The notion that the right hemisphere has some linguistic abilities is supported by the studies on split-brains \citep{sperry1961} and by the patterns of recovery of aphasic patients after lesions in the left hemisphere \citep{dronkers2017}.
Moreover, a number of brain imaging studies have confirmed the right hemisphere involvement in higher-level language tasks, such as comprehending metaphors or jokes, generating the best endings to sentences, mentally repairing grammatical errors, detecting story inconsistencies (see \citet{jungbeemanbilateral2005,beeman2013}).
All in all, this suggests that the right hemisphere is apt at recognizing distant relations between words. This conclusion is further reinforced by our observation of long-range (paragraph-level) context effects in the right hemisphere (Fig.\ref{Figure:Context}, Long).

The effects we observed in the right hemisphere are not simply the mirror image of the left hemisphere. Spatially, syntax and semantics dissociate more in the left than the right. (see Fig.\ref{Figure:BrainFit}, Panel B).
Moreover, the regions of overlap correspond to the regions integrating long context (Fig.\ref{Figure:Context}C, bottom row),  suggesting that the left hemisphere is relatively more involved in the processing of local semantic or syntactic information, whereas the right hemisphere integrates both information at a larger time-scale (supra sentential).

\subsection*{Syntax drives the integration of contextual information.}

The comparison between the predictions of the integral model trained on the intact texts, and the predictions of the combined syntactic and semantic embeddings from the information-restricted models (Fig.\ref{Figure:UniqueCorrelation}C), highlights a striking contrast between GloVe and GPT2. While the former, a purely lexical model, does not benefit from being trained on the integral text, GPT-2 shows clear synergetic effects of syntactic and semantic information. 
GPT-2's embeddings fit brain activation better when syntactic and semantic information can contribute together. The fact that the regions that benefit most from this synergetic effect are high-level integrative regions, at the end of the temporal processing hierarchy described by \cite{chang_information_2022}, suggests that the availability of syntactic information drives the semantic interpretation at the sentence level.

These regions are quite similar to the semantic peak regions highlighted in Fig.~\ref{Figure:BrainFit}A, and overlap with the regions showing context effects (Fig.\ref{Figure:Context}). This replicates, and extends, the results from \cite{jainincorporating2018} who, varying the amount of context fed to LSTM models, from 0 to 19 words, found shorter context effects in temporal regions (ibid. Fig 4).

\subsection*{Limitations of our study}

Two limitations of our study must be acknowledged.

The dissociation between syntax and semantics is not perfect. The way we created the semantic dataset by removing function words  clearly impacts supra-lexical semantics. For example, removing instances of \emph{and} and \emph{or} prevents the NLP model from distinguishing between the meaning of “A or B” and “A and B”. In other words, the logical form of sentences can be perturbed. This may partly explain the synergetic effect of syntax and semantics described above. Removing pronouns is also problematic as this removed the arguments of some verbs. Ideally, one would like to find transformations of the sentences that keep the semantic information associated to the function words like conjunctions or pronouns, but it is not clear how to do that.

A second limitation concerns potential confounding effects of prosody. One cannot exclude that the embeddings of the models captured some prosodic variables correlated with syntax \citep{bennett_syntaxprosody_2019}. For example, certain categories of words (e.g. determiners or pronouns) are shorter and less accented than others. Also, although the models are purely trained on written text, they acquire the capacity to predict the end of sentences, which are more likely to be followed by pauses in the acoustic signal. We included acoustic energy and the words' offsets in the baseline models to try and diminish the impact of such factors, but such controls cannot be perfect.  One way to address this issue would be to have participants \emph{read} the text, presented at a fixed presentation rate. This would effectively remove all low-level effects of prosody.

\subsection*{Conclusion}

State-of-the-art Natural Language Processing models, like transformers, trained with large enough corpora, can generate essentially flawless grammatical text, showing that they can acquire the grammar of the language.
Using them to fit brain data has become a common endeavour, even if their architecture rules them out of plausible models of the brain.
Yet, despite their low biological plausibility, their ability to build rich distributed representations can be exploited to study language processing in the brain.
In this paper, we have demonstrated that restricting information provided to the model during training can be used to show which brain areas encode this information.
Information-restricted models are powerful and flexible tools to probe the brain as they can be used to investigate whatever representational space chosen, such as semantics, syntax or context.
Moreover, once they are trained, these models can be used directly on any dataset in order to generate information-restricted features for model-brain alignment.
This approach is highly beneficial, both in term of richness of the features, and scalability, compared to classical approaches that use manually crafted features or focus on specific contrasts.
In future experiments, more fine grained control of both the information given to the models as well as model's representations will permit more precise characterisation of the role of the various regions involved in language comprehension.

\section{Methods and Materials}
\label{Methods}

\subsection{Creation of datasets; Semantic, Syntactic and Integral features} 
\label{Method:Datasets}

We selected a collection of English novels from Project Gutenberg (www.gutenberg.org; data retrieved on February 21, 2016). This \emph{original dataset} comprised 4.4GB of text for training purposes and 1.1GB for validation. From it, we created two information-restricted datasets: the \emph{semantic dataset} and the \emph{syntactic dataset}. In the \emph{semantic dataset}, only content words were kept, while all grammatical, function words and punctuation signs were filtered out. In the \emph{syntactic dataset}, each token (word or punctuation sign) was replaced by an identifier encoding a triplet (POS, Morph, NCN) where POS is the Part-of-speech computed using Spacy \citep{spacy2}, Morph corresponds to the morphological features obtained from Spacy and NCN stands for the Number of Closing Nodes in the parse tree, at the current token, computed using the Berkeley Neural Parser \citep{kitaevklein2018constituency} available with Spacy. 

In this paper, we refer to the content of the original dataset as \emph{integral features}, the content of the semantic dataset as \emph{semantic features}, and the content of the syntactic dataset as \emph{syntactic features}.
Examples of integral, semantic and syntactic features are given in Appendix 1-\nameref{Appendix:ModelTraining}.

\subsection{GloVe Training}
\label{Method:GloveTraining}

GloVe (Global Vectors for Word Representation) relies on the co-occurence matrix of words in a given corpus to generate fixed embedding vectors that capture the distributional properties of the words \citep{penningtonglove2014}. Using the open-source code provided by Pennington and al. (\url{https://nlp.stanford.edu/projects/glove/}), we trained GloVe on the three datasets (integral, semantic and syntactic), setting the context window size to 15 words, the embedding vectors' size to 768, and the number of training epochs to 23.

\subsection{GPT-2 Training}
\label{Method:GPT2Training}

GPT-2 (Generative Pretrained Transformer 2) is a deep learning transformer-based language model.
We trained the open-source implementation GPT2LMHeadModel, provided by HuggingFace \citep{wolf20}, on the three datasets (integral, semantic and syntactic).

The GPT2LMHeadModel architecture is trained on a next-token prediction task using a CrossEntropyLoss and the Pytorch python package \citep{NEURIPS20199015}.
The training procedure can easily be extended to any feature type by adapting both vocabulary size and tokenizer to each vocabulary.
Indeed, the inputs given to GPT2LMHeadModel are ids encoding vocabulary items.
All the analyses reported in this paper were performed with 4-layer models having 768 units per layer and 12 attention heads. 
As shown in \citep{pasquiouicml2022}, these 4-layer models fit brain data nearly as well as the usual 12-layer models. 
We presented the models with input sequences of 512 tokens, and let the training run for 5 epochs; convergence assessments are provided in Appendix 1-\nameref{Appendix:ModelConvergence} (Appendix 1-Fig.\ref{Appendix:Figure:ModelConvergence}).

For the GTP-2 trained on the semantic features, small modifications had to be made to the model architecture in order to remove all residual syntax.
By default, GPT-2 encodes the absolute positions of tokens in sentences. 
When training GPT-2 on the semantic features, as word ordering might contain syntactic information, we had to make sure that position information could not be leveraged by means of its positional embeddings, yet keeping information about word proximity as it influences semantics. We modified the implementation so that the GPT-2 trained on semantic features follows these specifications (see Appendix 1-\nameref{Appendix:PositionalEmbeddings}).

\subsection{Stimulus: The Little Prince story}
\label{Method:Stimulus}

The stimulus used to obtain activations from humans and from NLP models was \emph{The Little Prince} novella. Humans listened to an audio-book version, spliced into 9 tracks that lasted approximately 11 minutes each \citep[see][]{LPPdatapaper}. In parallel, NLP models were provided with an exact transcription of this audio-book, enriched with punctuation signs from the written version of the Little Prince. The text comprised 15,426 words and 4,482 punctuation signs. The acoustic onsets and offsets of the spoken words were marked to align the audio recording with the \emph{The Little Prince} text. 

\subsection{Computing Embeddings from the Little Prince text}
\label{Method:ComputingEmbeddings}

The tokenized versions of the Little Prince (one for each feature type) were run through Glove and GPT-2 in order to generate embeddings that could be compared with fMRI data.

For GloVe, we simply retrieved the fixed embedding vector learnt during training for each token.

For GPT-2, we retrieved the contextualized third layer hidden-state (aka embedding) vector for each token, so that the dimension is comparable to the dimension of GloVe's embeddings (768 units).
Layer 3 (out of 4) was selected because it has been demonstrated that late middle layers of recurrent language models are best able to predict brain activity \citep{tonevainterpreting2019,jainincorporating2018}.

The embedding built by GPT-2 for a given token rely on the past tokens (aka past context). The bigger the past context, the more reliable the token embedding will be. We designed the following procedure to ensure that the embedding of each token used similar past context size:
the input sequence was limited to a maximum of 512 tokens. 
The text was scanned with a sliding window of size $N=512$ tokens, and a step of 1 token. The embedding vector of the next to last token (in the sliding window) was then retrieved.
For the context-constrained versions of GPT-2 (denoted GPT-$2_{Context-k}$), the input text was formatted as the training data (see Fig.\ref{Fig:Pipeline}C) in batches of input sequences of length ($k+5$) tokens (see Appendix 1-\nameref{Appendix:ContextLimitedInput} for examples), and only the embedding vector of the current token was retrieved.
Embedding matrices were built by concatenating words embeddings.
More precisely, calling $d$ the dimension of the embeddings retrieved from of a neural model, corresponding to the number of units in one layer in our case, and $w$ the total number of tokens in the text, we obtained an embedding matrix $\mathbf{X} \in \mathbb{R}^{w \times d}$ after the presentation of the entire text to the model. 

\subsection{Decoding of syntax and semantics categories from embeddings}
\label{Method:Decoding}

We designed two decoding tasks: a syntax decoding task in which we tried to predict the triplet (Part-of-speech, morphological information and number of closing nodes) of each word from its embedding vector (355 categories), and a semantic decoding task in which we tried to predict each word's semantic category (from \textit{Wordnet}, \url{https://wordnet.princeton.edu/}) from its embedding vector (837 categories).

We used Logistic Classifiers and the text of \emph{The Little Prince} as train and test data, which was split using a 9-fold cross-validation on runs, training on 8 runs and evaluating on the remaining one for each split. Dummy classifiers were fitted and used as estimations of chance-level for each task and model. 
All classifiers implementations were taken from Scikit-Learn \citep{scikitlearn}.

\subsection{MRI data}
\label{Method:MRIData}

We used the functional Magnetic Resonance Imaging (fMRI) data of 51 English speaking participants who listened to an entire audio-book of The Little Prince during about one hour and a half. 
These data, available at \url{https://openneuro.org/datasets/ds003643/versions/1.0.2} are described in details by \cite{LPPdatapaper}. 
In short, the acquisition used echo-planar imaging (TR=2s; resolution=3.75x3.75x3.75mm) with a multi-echo (3 echos) sequence to optimize signal-to-noise \citep{kundumultiecho2017}. 
Preprocessing comprised multi-echo independent components analysis (ME-ICA) to denoise data for motion, physiology and scanner artifacts, correction for slice-timing differences, and nonlinear alignment to the MNI template brain.

For each participant, there were 9 runs of fMRI acquisition representing about 10 minutes of brain activations each.
We re-sampled the preprocessed individual scans at 4x4x4 mm (to reduce computation load) and applied linear detrending and standardization (mean removal and scaling to unit variance) to each voxel's time-series.

Finally, we computed a global brain mask to only keep voxels containing useful signal (using nilearn's \textit{compute\_epi\_mask} function, we find the least dense point of the total image histogram) across all runs for at least 50\% of all participants. This global mask contained 26,164 voxels at 4x4x4mm resolution. All analyses reported in this paper were performed within this global mask.

\subsection{Correlations between embeddings and individual fMRI data}
\label{Method:EncodingParadigm}

The embeddings ($\textbf{X}$) derived from neural language models were mapped to each subject's fMRI activations  ($\textbf{Y}_{s}, s=1..S$) following the pipeline outlined in Fig.\ref{Fig:Pipeline}B.

The process, using the standard model-based encoding approach to modelling fMRI signals \citep{huthnatural2016,naselarisencoding2011,pasquiouicml2022}, is detailed in Appendix 1-\nameref{Appendix:MappingNlmToBrain}. 
In brief, each column of $\textbf{X}$ was first aligned with the words' offsets in the audio stream and convolved with the default \textit{SPM} haemodynamic kernel (using Nilearn's \textit{compute\_regressor} function from the \textit{nilearn.glm.first\_level} module).
The resulting time-course was sub-sampled to match the sampling frequency of the scans $\textbf{Y}_{s}$ (giving $\tilde{\textbf{X}}$). 
Next, in each individual voxel, the time-course of brain activation was regressed on $\tilde{\textbf{X}}$ using Ridge regression.
The Ridge regression regularization was estimated using a nested-cross validation scheme (see Appendix 1-\nameref{Appendix:MappingNlmToBrain} for more details).
Finally, the cross-validated Pearson correlation $R$ between the encoding model's prediction and the fMRI signal for subject $s$ in voxel $v$ was computed. 
The output of this process is a map of correlations between the encoding models' predictions and the observed time series, for a given participant.

\subsection{Baseline fMRI model}
\label{Method:BF}

To obtain a more accurate evaluation of the specific impact of the embeddings on brain scores, we removed the contribution of three confounding variables from all maps presented in this paper. The three confounding variables were: a) \emph{the acoustic energy} (root mean squared of the audio signal sampled every 10ms) b) \emph{the word-rate} (one event at each word offset) c) \emph{the log of the unigram lexical frequency} of each word (modulator of the word events). An fMRI Ridge linear model that only included these three regressors was used to compute a map of cross-validated correlations for each participant.

The $R$-maps presented in Fig.\ref{Figure:BrainFit} of this paper are corrected for the contribution of these variables, that is they display $\Delta R$, the increase in $R$ when adding a model to the baseline model versus the baseline model by itself.

Appendix 1-Fig.\ref{Appendix:Figure:BF} displays the significant correlations in the group-level $R$ maps associated with the Baseline Model, corrected for multiple comparison using a FDR correction ($p<0.005$).

\subsection{Group-level Maps}
\label{Method:GroupAnalyses}

The brain maps presented in this document display group average increase in $R$ scores obtained from individuals correlation maps (relative to the baseline model or to another model).  
Only voxels showing statistically significant increase in $R$ score are shown.

Significance was assessed through one-sample t-tests applied to the spatially smoothed correlation maps, with an isotropic Gaussian kernel with FWHM of 6mm. 
In each voxel, the test assessed whether the distribution of $R_{test}$ values across participants was significantly larger than zero.
To control for multiple comparisons, all maps were corrected using a False Discovery Rate (FDR) correction with $p<0.005$ \citep{benjaminiFDR}. On each corrected figure, the FDR threshold on the z-scores, named $z_{FDR}$, is indicated at the bottom, that is, values reported on these maps (e.g. $R$ scores) are shown only for voxels whose z-score survived this threshold ($z_{voxel}>z_{FDR}$).

While all analyses were done on volume data, all brain maps were projected onto brain surface for visualization purposes, using ‘\textit{fsaverage5}' (from Nilearn's \textit{datasets.fetch\_surf\_fsaverage}) mesh and the ‘\textit{vol\_to\_surf}' function (from Nilearn's \textit{surface} module).

\subsection{Syntax and Semantics peak regions}
\label{Method:PeakRegions}

We decided to also report brain maps' \textit{peak regions}, i.e. the 10\% of the voxels having the highest $R$ score in a brain map.
The motivation is that two different language processes might elicit lots of brain regions in common, while the regions that are better fitted by the representations derived from each process might differ. 
The peak regions of the neural correlates of semantic and syntactic representations are displayed on surface brain maps.
The proportions of voxels belonging to each peak region as well as the Jaccard score between syntax and semantics are displayed for each model and hemisphere.

\subsection{Jaccard index}
\label{Method:Jaccard}

The Jaccard index (computed using scikit-learn \textit{jaccard\_score} function from the \textit{metrics} module) for two sets $X$ and $Y$ is defined in the following manner: $J(X,Y) = |X \cap Y| / |X \cup Y|$. It behaves as a similarity coefficient: when the two sets completely overlap, J=1; when their intersection  is nil, J=0.

\subsection{Specificity index}
\label{Method:SpecificityIndex}

To quantify how much each voxel $v$ is influenced by semantic and syntactic embeddings, we defined a \textit{specificity index} in the following manner:

$$
x_{sensitivity}(v) = \log_{10}\left(\frac{r_{Semantic}(v)}{r_{Syntax}(v)}\right)
$$

$r_{Syntax}$ is the $R$ score increase relative to the baseline model for the syntactic embeddings.
$r_{Semantic}$ is the $R$ score increase relative to the baseline model for the semantic embeddings.

In Fig.\ref{Figure:DiffRatio}, the higher and greener $x$ is, the more sensitive it is to semantic embeddings compared to syntactic embeddings.
The lower and redder $x$ is, the more sensitive it is to syntactic embeddings compared to semantic embeddings.
$x$ close to 0, indicates an equal sensitivity to syntactic and semantic embeddings. 

Group average specificity index maps were computed from each subject's map and significance was assessed through one-sample t-tests applied to the spatially smoothed specificity maps, with an isotropic Gaussian kernel with FWHM of 6mm. 
A FDR correction ($p<0.005$) was used to correct for multiple comparisons.

\subsection{Meta-Analysis based on Neurosynth}
\label{Methods:MetaAnalysis}

We used the \textit{Neurosynth} database (\url{https://github.com/neurosynth/neurosynth}) to perform a meta-analysis of brain regions that appeared in fMRI articles containing the words 'syntactic' or 'semantic' in their abstract. 
Using a frequency threshold of 0.05, the keyword \emph{semantic} yielded 626 articles, while \textit{syntactic} yielded 128 articles. 

The \textit{meta.MetaAnalysis} function from the neurosynth package was then used to create association test maps for syntax and semantics.
These maps display voxels that are reported more often in articles that mention the keyword than articles that do not. Such association test maps indicate whether or not there’s a non-zero association between activation of the voxel in question and the use of a particular term in a study. 
We fused the maps associated to \emph{syntactic} and \emph{semantic}, thresholded with a False Discovery Rate set to 0.01, to produce Fig.\ref{Figure:MetaAnalysis}.

\section{Data Availability}
\label{DataAvailability}

The Integral Dataset (train, test and dev) is available at: \url{https://osf.io/jzcvu/}.
The semantic and syntactic datasets can be derived from the Integral Dataset using the scripts provided in \url{https://github.com/AlexandrePsq/Information-Restrited-NLMs}.

All analyses, as well as model training, features extraction and the fitting of encoding models were performed using Python 3.7.6 and can be replicated using the code provided in the same Github repository (\url{https://github.com/AlexandrePsq/Information-Restrited-NLMs}). The required packages are listed there. A non-exhaustive list includes numpy \citep{Harris2020}, scipy \citep{Virtanen2020}, scikit-learn \citep{scikitlearn}, matplotlib \citep{matplotlib}, pandas \citep{mckinney2010data} and nilearn (\url{https://nilearn.github.io/stable/index.html}).

The fMRI dataset is publicly available at \url{https://openneuro.org/datasets/ds003643/versions/1.0.2}, and all details regarding the dataset are described in details by \cite{LPPdatapaper}.

\section{Acknowledgments}

This project/research has received funding from the American National Science Foundation under Grant Number 1607441 (USA), the French National Research Agency (ANR) under grant ANR-14-CERA-0001, the European Union’s Horizon 2020 Framework Programme for Research and Innovation under the Specific Grant Agreement No. 945539 (Human Brain Project SGA3), and the KARAIB AI chair (ANR-20-CHIA-0025-01).

\bibliography{ms}

\begin{thebibliography}{93}
\providecommand{\natexlab}[1]{#1}
\providecommand{\urlprefix}{}
\providecommand{\doiprefix}{doi: }

\bibitem[{Baetens et~al.(2014)Baetens, Kris and Ma, Ning and Steen, Johan and
  Van Overwalle, Frank}]{baetens_involvement_2014}
\textbf{\color{eLifeMediumGrey} Baetens K}, Ma N, Steen J, Van~Overwalle F.
\newblock Involvement of the mentalizing network in social and non-social high
  construal.
\newblock Social Cognitive and Affective Neuroscience.  2014 Jun;
  9(6):817--824.
\newblock \urlprefix\url{https://doi.org/10.1093/scan/nst048},
  \href{10.1093/scan/nst048}{\doiprefix \detokenize{10.1093/scan/nst048}}.

\bibitem[{Baldassano et~al.(2017)Christopher Baldassano and Janice Chen and
  Asieh Zadbood and Jonathan W. Pillow and Uri Hasson and Kenneth A.
  Norman}]{BALDASSANO2017709}
\textbf{\color{eLifeMediumGrey} Baldassano C}, Chen J, Zadbood A, Pillow JW,
  Hasson U, Norman KA.
\newblock Discovering Event Structure in Continuous Narrative Perception and
  Memory.
\newblock Neuron.  2017; 95(3):709--721.e5.
\newblock
  \urlprefix\url{https://www.sciencedirect.com/science/article/pii/S0896627317305937},
  \href{https://doi.org/10.1016/j.neuron.2017.06.041}{\doiprefix
  \detokenize{https://doi.org/10.1016/j.neuron.2017.06.041}}.

\bibitem[{Bates and Dick(2002)Bates, Elizabeth and Dick,
  Frederic}]{bates2002language}
\textbf{\color{eLifeMediumGrey} Bates E}, Dick F.
\newblock Language, gesture, and the developing brain.
\newblock Developmental Psychobiology: The Journal of the International Society
  for Developmental Psychobiology.  2002; 40(3):293--310.
\newblock \urlprefix\url{https://pubmed.ncbi.nlm.nih.gov/11891640/}, publisher:
  Wiley Online Library.

\bibitem[{Bates and MacWhinney(1989)Elizabeth Bates and Brian
  MacWhinney}]{bates1989functionalism}
\textbf{\color{eLifeMediumGrey} Bates E}, MacWhinney B.
\newblock Functionalism and the Competition Model.
\newblock In: MacWhinney B, Bates E, editors. \emph{The Crosslinguistic Study
  of Sentence Processing} Cambridge University Press; 1989.p. 3--73.
\newblock
  \urlprefix\url{https://www.researchgate.net/publication/230875840_Functionalism_and_the_Competition_Model/link/545a97170cf2c16efbbbc1d5/download}.

\bibitem[{Beeman and Chiarello(2013)Beeman, Mark Jung and Chiarello,
  Christine}]{beeman2013}
\textbf{\color{eLifeMediumGrey} Beeman MJ}, Chiarello C.
\newblock Right hemisphere language comprehension: {Perspectives} from
  cognitive neuroscience.
\newblock Psychology Press; 2013.
\newblock
  \urlprefix\url{https://www.taylorfrancis.com/books/mono/10.4324/9780203763544/right-hemisphere-language-comprehension-mark-jung-beeman-christine-chiarello}.

\bibitem[{Benjamini and Hochberg(1995)Yoav Benjamini and Yosef
  Hochberg}]{benjaminiFDR}
\textbf{\color{eLifeMediumGrey} Benjamini Y}, Hochberg Y.
\newblock Controlling the False Discovery Rate: A Practical and Powerful
  Approach to Multiple Testing.
\newblock Journal of the Royal Statistical Society Series B (Methodological).
  1995; 57(1):289--300.
\newblock \urlprefix\url{http://www.jstor.org/stable/2346101}.

\bibitem[{Bennett and Elfner(2019)Bennett, Ryan and Elfner,
  Emily}]{bennett_syntaxprosody_2019}
\textbf{\color{eLifeMediumGrey} Bennett R}, Elfner E.
\newblock The {Syntax}–{Prosody} {Interface}.
\newblock Annual Review of Linguistics.  2019 Jan; 5(1):151--171.
\newblock
  \urlprefix\url{https://www.annualreviews.org/doi/10.1146/annurev-linguistics-011718-012503},
  \href{10.1146/annurev-linguistics-011718-012503}{\doiprefix
  \detokenize{10.1146/annurev-linguistics-011718-012503}}.

\bibitem[{Binder and Desai(2011)Binder, Jeffrey R. and Desai, Rutvik
  H.}]{binder_neurobiology_2011}
\textbf{\color{eLifeMediumGrey} Binder JR}, Desai RH.
\newblock The neurobiology of semantic memory.
\newblock Trends in Cognitive Sciences.  2011 Nov; 15(11):527--536.
\newblock
  \urlprefix\url{http://linkinghub.elsevier.com/retrieve/pii/S1364661311002142},
  \href{10.1016/j.tics.2011.10.001}{\doiprefix
  \detokenize{10.1016/j.tics.2011.10.001}}.

\bibitem[{Binder et~al.(2009)Binder, Jeffrey R. and Desai, Rutvik H. and
  Graves, William W. and Conant, Lisa L.}]{binder2009review}
\textbf{\color{eLifeMediumGrey} Binder JR}, Desai RH, Graves WW, Conant LL.
\newblock {Where Is the Semantic System? A Critical Review and Meta-Analysis of
  120 Functional Neuroimaging Studies}.
\newblock Cerebral Cortex.  2009 03; 19(12):2767--2796.
\newblock \urlprefix\url{https://doi.org/10.1093/cercor/bhp055},
  \href{10.1093/cercor/bhp055}{\doiprefix \detokenize{10.1093/cercor/bhp055}}.

\bibitem[{Bottini et~al.(1995)Bottini, Gabriella and Corcoran, Rhiannon and
  Sterzi, Roberto and Paulesu, Eraldo and Schenone, Pietro and Scarpa, Pina and
  Frackowiak, Richard and Frith, Chris}]{bottini1995righthemisphere}
\textbf{\color{eLifeMediumGrey} Bottini G}, Corcoran R, Sterzi R, Paulesu E,
  Schenone P, Scarpa P, Frackowiak R, Frith C.
\newblock The role of the right hemisphere in the interpretation of figurative
  aspects of language. A positron emission tomography activation study.
\newblock Brain : a journal of neurology.  1995 01; 117 ( Pt 6):1241--53.
\newblock
  \urlprefix\url{https://www.researchgate.net/publication/15377772_The_role_of_the_right_hemisphere_in_the_interpretation_of_figurative_aspects_of_language_A_positron_emission_tomography_activation_study},
  \href{10.1093/brain/117.6.1241}{\doiprefix
  \detokenize{10.1093/brain/117.6.1241}}.

\bibitem[{Caplan et~al.(1998)Caplan, David and Alpert, Nathaniel and Waters,
  Gloria}]{caplaneffects1998}
\textbf{\color{eLifeMediumGrey} Caplan D}, Alpert N, Waters G.
\newblock Effects of {Syntactic} {Structure} and {Propositional} {Number} on
  {Patterns} of {Regional} {Cerebral} {Blood} {Flow}.
\newblock Journal of Cognitive Neuroscience.  1998 Jul; 10(4):541--552.
\newblock \urlprefix\url{https://doi.org/10.1162/089892998562843},
  \href{10.1162/089892998562843}{\doiprefix
  \detokenize{10.1162/089892998562843}}, \_eprint:
  https://direct.mit.edu/jocn/article-pdf/10/4/541/1931814/089892998562843.pdf.

\bibitem[{Caramazza and Zurif(1976)Caramazza, Alfonso and Zurif, Edgar
  B}]{caramazza1976dissociation}
\textbf{\color{eLifeMediumGrey} Caramazza A}, Zurif EB.
\newblock Dissociation of algorithmic and heuristic processes in language
  comprehension: Evidence from aphasia.
\newblock Brain and language.  1976; 3(4):572--582.
\newblock \urlprefix\url{https://pubmed.ncbi.nlm.nih.gov/974731/}.

\bibitem[{Caucheteux et~al.(2021)Caucheteux, Charlotte and Gramfort, Alexandre
  and King, Jean-Remi}]{caucheteux:icml2021}
\textbf{\color{eLifeMediumGrey} Caucheteux C}, Gramfort A, King JR.
\newblock {Disentangling Syntax and Semantics in the Brain with Deep Networks}.
\newblock In: \emph{{ICML 2021 - 38th International Conference on Machine
  Learning}} Online conference, France; 2021. p.~13.
\newblock \urlprefix\url{https://hal.archives-ouvertes.fr/hal-03361421}.

\bibitem[{Caucheteux and King(2022)Caucheteux, Charlotte and King,
  Jean-R{\'e}mi}]{Caucheteux:2022}
\textbf{\color{eLifeMediumGrey} Caucheteux C}, King JR.
\newblock Brains and algorithms partially converge in natural language
  processing.
\newblock Communications Biology.  2022;
  \urlprefix\url{https://pubmed.ncbi.nlm.nih.gov/35173264/},
  \href{10.1038/s42003-022-03036-1}{\doiprefix
  \detokenize{10.1038/s42003-022-03036-1}}.

\bibitem[{Chang et~al.(2022)Chang, Claire H. C. and Nastase, Samuel A. and
  Hasson, Uri}]{chang_information_2022}
\textbf{\color{eLifeMediumGrey} Chang CHC}, Nastase SA, Hasson U.
\newblock Information flow across the cortical timescale hierarchy during
  narrative construction.
\newblock Proceedings of the National Academy of Sciences.  2022 Dec;
  119(51):e2209307119.
\newblock \urlprefix\url{http://www.pnas.org/doi/full/10.1073/pnas.2209307119},
  \href{10.1073/pnas.2209307119}{\doiprefix
  \detokenize{10.1073/pnas.2209307119}}, publisher: Proceedings of the National
  Academy of Sciences.

\bibitem[{Chomsky(1984)Chomsky, Noam}]{chomskymodular1984}
\textbf{\color{eLifeMediumGrey} Chomsky N}.
\newblock Modular {Approaches} to the {Study} of the {Mind}, vol.~1.
\newblock San Diego State University Press San Diego; 1984.
\newblock
  \urlprefix\url{https://archive.org/details/modularapproache00noam/page/n9/mode/2up}.

\bibitem[{Cooke et~al.(2001)Cooke, Ayanna and Zurif, Edgar B. and DeVita,
  Christian and Alsop, David and Koenig, Phyllis and Detre, John and Gee, James
  and PinÃ£ngo, Maria and Balogh, Jennifer and Grossman,
  Murray}]{cookeneural2001}
\textbf{\color{eLifeMediumGrey} Cooke A}, Zurif EB, DeVita C, Alsop D, Koenig
  P, Detre J, Gee J, PinÃ£ngo M, Balogh J, Grossman M.
\newblock Neural basis for sentence comprehension: {Grammatical} and short term
  memory components.
\newblock Human Brain Mapping.  2001 Nov; 15(2):80--94.
\newblock
  \urlprefix\url{https://www.ncbi.nlm.nih.gov/pmc/articles/PMC6872024/},
  \href{10.1002/hbm.10006}{\doiprefix \detokenize{10.1002/hbm.10006}}.

\bibitem[{Damasio et~al.(2004)Damasio, H. and Tranel, D. and Grabowski, T. and
  Adolphs, R. and Damasio, A.}]{damasio2004}
\textbf{\color{eLifeMediumGrey} Damasio H}, Tranel D, Grabowski T, Adolphs R,
  Damasio A.
\newblock Neural systems behind word and concept retrieval.
\newblock Cognition.  2004; 92(1-2):179--229.
\newblock \href{10.1016/j.cognition.2002.07.001}{\doiprefix
  \detokenize{10.1016/j.cognition.2002.07.001}}.

\bibitem[{Devlin et~al.(2019)Devlin, Jacob and Chang, Ming-Wei and Lee, Kenton
  and Toutanova, Kristina}]{devlinbert2019}
\textbf{\color{eLifeMediumGrey} Devlin J}, Chang MW, Lee K, Toutanova K.
\newblock {BERT}: {Pre}-training of {Deep} {Bidirectional} {Transformers} for
  {Language} {Understanding}.
\newblock arXiv:181004805 [cs].  2019 May;
  \urlprefix\url{http://arxiv.org/abs/1810.04805}, arXiv: 1810.04805.

\bibitem[{Dick et~al.(2001)Dick, Frederic and Bates, Elizabeth and Wulfeck,
  Beverly and Utman, Jennifer Aydelott and Dronkers, Nina and Gernsbacher,
  Morton Ann}]{dicklanguage2001}
\textbf{\color{eLifeMediumGrey} Dick F}, Bates E, Wulfeck B, Utman JA, Dronkers
  N, Gernsbacher MA.
\newblock Language deficits, localization, and grammar: evidence for a
  distributive model of language breakdown in aphasic patients and
  neurologically intact individuals.
\newblock Psychological review.  2001; 108(4):759.
\newblock \urlprefix\url{https://psycnet.apa.org/record/2001-18918-004},
  publisher: American Psychological Association.

\bibitem[{Dronkers et~al.(2017)Dronkers, Nina F. and Ivanova, Maria V. and
  Baldo, Juliana V.}]{dronkers2017}
\textbf{\color{eLifeMediumGrey} Dronkers NF}, Ivanova MV, Baldo JV.
\newblock What {Do} {Language} {Disorders} {Reveal} about {Brain}–{Language}
  {Relationships}? {From} {Classic} {Models} to {Network} {Approaches}.
\newblock Journal of the International Neuropsychological Society : JINS.  2017
  Oct; 23(9-10):741--754.
\newblock
  \urlprefix\url{https://www.ncbi.nlm.nih.gov/pmc/articles/PMC6606454/},
  \href{10.1017/S1355617717001126}{\doiprefix
  \detokenize{10.1017/S1355617717001126}}.

\bibitem[{Elman(1991)Elman, Jeffrey}]{elmandistributed1991}
\textbf{\color{eLifeMediumGrey} Elman J}.
\newblock Distributed representations, simple recurrent networks, and
  grammatical structure.
\newblock Machine Learning.  1991; 7:195--225.
\newblock \urlprefix\url{https://link.springer.com/article/10.1007/BF00114844}.

\bibitem[{Embick(2000)David Embick}]{embick2000syntactic}
\textbf{\color{eLifeMediumGrey} Embick D}.
\newblock Features, Syntax, and Categories in the Latin Perfect.
\newblock Linguistic Inquiry.  2000; 31(2):185--230.
\newblock \urlprefix\url{http://www.jstor.org/stable/4179104}.

\bibitem[{Fedorenko et~al.(2020)Fedorenko, Evelina and Blank, Idan and
  Siegelman, Matthew and Mineroff, Zachary}]{fedorenkolack2020}
\textbf{\color{eLifeMediumGrey} Fedorenko E}, Blank I, Siegelman M, Mineroff Z.
\newblock Lack of selectivity for syntax relative to word meanings throughout
  the language network.
\newblock bioRxiv.  2020; p. 477851.
\newblock
  \urlprefix\url{https://www.sciencedirect.com/science/article/pii/S0010027720301670},
  publisher: Cold Spring Harbor Laboratory.

\bibitem[{Ferstl and von Cramon(2001)Ferstl, Evelyn C. and von Cramon, D.
  Yves}]{ferstl_role_2001}
\textbf{\color{eLifeMediumGrey} Ferstl EC}, von Cramon DY.
\newblock The role of coherence and cohesion in text comprehension: an
  event-related {fMRI} study.
\newblock Cognitive Brain Research.  2001 Jun; 11(3):325--340.
\newblock
  \urlprefix\url{http://www.sciencedirect.com/science/article/pii/S0926641001000076},
  \href{10.1016/S0926-6410(01)00007-6}{\doiprefix
  \detokenize{10.1016/S0926-6410(01)00007-6}}.

\bibitem[{Fodor(1983)Fodor, Jerry}]{fodormodularity1983}
\textbf{\color{eLifeMediumGrey} Fodor J}.
\newblock The modularity of mind.
\newblock MIT press; 1983.
\newblock
  \urlprefix\url{https://mitpress.mit.edu/9780262560252/the-modularity-of-mind/}.

\bibitem[{Friederici(2011)Friederici, Angela D}]{friedericibrain2011}
\textbf{\color{eLifeMediumGrey} Friederici AD}.
\newblock The {Brain} {Basis} of {Language} {Processing}: {From} {Structure} to
  {Function}.
\newblock Physiol Rev.  2011; 91:36.
\newblock \urlprefix\url{https://pubmed.ncbi.nlm.nih.gov/22013214/}.

\bibitem[{Friederici et~al.(2017)Friederici, Angela D and Chomsky, Noam and
  Berwick, Robert C and Moro, Andrea and Bolhuis, Johan
  J}]{friederici2017language}
\textbf{\color{eLifeMediumGrey} Friederici AD}, Chomsky N, Berwick RC, Moro A,
  Bolhuis JJ.
\newblock Language, mind and brain.
\newblock Nature human behaviour.  2017; 1(10):713--722.

\bibitem[{Friederici et~al.(2006)Friederici, Angela D. and Fiebach, Christian
  J. and Schlesewsky, Matthias and Bornkessel, Ina D. and von Cramon, D.
  Yves}]{friederici2006processing}
\textbf{\color{eLifeMediumGrey} Friederici AD}, Fiebach CJ, Schlesewsky M,
  Bornkessel ID, von Cramon DY.
\newblock {Processing Linguistic Complexity and Grammaticality in the Left
  Frontal Cortex}.
\newblock Cerebral Cortex.  2006 01; 16(12):1709--1717.
\newblock \urlprefix\url{https://doi.org/10.1093/cercor/bhj106},
  \href{10.1093/cercor/bhj106}{\doiprefix \detokenize{10.1093/cercor/bhj106}}.

\bibitem[{Friederici et~al.(2009{\natexlab{a}})Friederici, Angela D. and Kotz,
  Sonja A. and Scott, Sophie K. and Obleser,
  Jonas}]{friedericidisentangling2009}
\textbf{\color{eLifeMediumGrey} Friederici AD}, Kotz SA, Scott SK, Obleser J.
\newblock Disentangling syntax and intelligibility in auditory language
  comprehension.
\newblock Human Brain Mapping.  2009 Aug; 31(3):448--457.
\newblock
  \urlprefix\url{https://www.ncbi.nlm.nih.gov/pmc/articles/PMC6870868/},
  \href{10.1002/hbm.20878}{\doiprefix \detokenize{10.1002/hbm.20878}}.

\bibitem[{Friederici et~al.(2009{\natexlab{b}})Friederici, Angela D. and
  Makuuchi, Michiru and Bahlmann, JÃ¶rg}]{friedericirole2009}
\textbf{\color{eLifeMediumGrey} Friederici AD}, Makuuchi M, Bahlmann J.
\newblock The role of the posterior superior temporal cortex in sentence
  comprehension.
\newblock NeuroReport.  2009 Apr; 20(6):563--568.
\newblock
  \urlprefix\url{https://journals.lww.com/neuroreport/Fulltext/2009/04220/The_role_of_the_posterior_superior_temporal_cortex.6.aspx},
  \href{10.1097/WNR.0b013e3283297dee}{\doiprefix
  \detokenize{10.1097/WNR.0b013e3283297dee}}.

\bibitem[{Friederici et~al.(2003)Friederici, Angela D. and RÃ¼schemeyer,
  Shirley-Ann and Hahne, Anja and Fiebach, Christian
  J.}]{friederici2003roleleftifg}
\textbf{\color{eLifeMediumGrey} Friederici AD}, RÃ¼schemeyer SA, Hahne A,
  Fiebach CJ.
\newblock {The Role of Left Inferior Frontal and Superior Temporal Cortex in
  Sentence Comprehension: Localizing Syntactic and Semantic Processes}.
\newblock Cerebral Cortex.  2003 02; 13(2):170--177.
\newblock \urlprefix\url{https://doi.org/10.1093/cercor/13.2.170},
  \href{10.1093/cercor/13.2.170}{\doiprefix
  \detokenize{10.1093/cercor/13.2.170}}.

\bibitem[{Friederici(2017)Friederici, Angela
  Dorkas}]{friedericineurobiology2017}
\textbf{\color{eLifeMediumGrey} Friederici AD}.
\newblock Neurobiology of {Syntax} as the {Core} of {Human} {Language}.
\newblock BIOLINGUISTICS.  2017; 11.
\newblock
  \urlprefix\url{https://bioling.psychopen.eu/index.php/bioling/article/view/9093}.

\bibitem[{Garrard et~al.(2004)P. Garrard and E. Carroll and D. Vinson and G.
  Vigliocco}]{garrard2004dissociation}
\textbf{\color{eLifeMediumGrey} Garrard P}, Carroll E, Vinson D, Vigliocco G.
\newblock Dissociation of Lexical Syntax and Semantics: Evidence from Focal
  Cortical Degeneration.
\newblock Neurocase.  2004; 10(5):353--362.
\newblock \urlprefix\url{https://doi.org/10.1080/13554790490892248},
  \href{10.1080/13554790490892248}{\doiprefix
  \detokenize{10.1080/13554790490892248}}, pMID: 15788273.

\bibitem[{Goodglass(1993)Goodglass, Harold}]{goodglass1993understanding}
\textbf{\color{eLifeMediumGrey} Goodglass H}.
\newblock Understanding aphasia.
\newblock Academic Press; 1993.
\newblock \urlprefix\url{https://www.jstor.org/stable/416147}.

\bibitem[{Grodzinsky and Santi(2008)Yosef Grodzinsky and Andrea
  Santi}]{GRODZINSKY2008474}
\textbf{\color{eLifeMediumGrey} Grodzinsky Y}, Santi A.
\newblock The battle for Broca's region.
\newblock Trends in Cognitive Sciences.  2008; 12(12):474--480.
\newblock
  \urlprefix\url{https://www.sciencedirect.com/science/article/pii/S1364661308002222},
  \href{https://doi.org/10.1016/j.tics.2008.09.001}{\doiprefix
  \detokenize{https://doi.org/10.1016/j.tics.2008.09.001}}.

\bibitem[{Hagoort(2014)Hagoort, Peter}]{hagoortnodes2014}
\textbf{\color{eLifeMediumGrey} Hagoort P}.
\newblock Nodes and networks in the neural architecture for language: {Broca}'s
  region and beyond.
\newblock Current opinion in Neurobiology.  2014; 28:136--141.
\newblock \urlprefix\url{https://pubmed.ncbi.nlm.nih.gov/25062474/}, publisher:
  Elsevier.

\bibitem[{Harris et~al.(2020)Harris, Charles R. and Millman, K. Jarrod and van
  der Walt, St{\'e}fan J. and Gommers, Ralf and Virtanen, Pauli and Cournapeau,
  David and Wieser, Eric and Taylor, Julian and Berg, Sebastian and Smith,
  Nathaniel J. and Kern, Robert and Picus, Matti and Hoyer, Stephan and van
  Kerkwijk, Marten H. and Brett, Matthew and Haldane, Allan and del R{\'i}o,
  Jaime Fern{\'a}ndez and Wiebe, Mark and Peterson, Pearu and
  G{\'e}rard-Marchant, Pierre and Sheppard, Kevin and Reddy, Tyler and
  Weckesser, Warren and Abbasi, Hameer and Gohlke, Christoph and Oliphant,
  Travis E.}]{Harris2020}
\textbf{\color{eLifeMediumGrey} Harris CR}, Millman KJ, van~der Walt SJ,
  Gommers R, Virtanen P, Cournapeau D, Wieser E, Taylor J, Berg S, Smith NJ,
  Kern R, Picus M, Hoyer S, van Kerkwijk MH, Brett M, Haldane A, del R{\'i}o
  JF, Wiebe M, Peterson P, G{\'e}rard-Marchant P, et~al.
\newblock Array programming with NumPy.
\newblock Nature.  2020 Sep; 585(7825):357--362.
\newblock \urlprefix\url{https://doi.org/10.1038/s41586-020-2649-2},
  \href{10.1038/s41586-020-2649-2}{\doiprefix
  \detokenize{10.1038/s41586-020-2649-2}}.

\bibitem[{Hashimoto and Sakai(2002)Ryuichiro Hashimoto and Kuniyoshi L
  Sakai}]{hashimoto2002specialization}
\textbf{\color{eLifeMediumGrey} Hashimoto R}, Sakai KL.
\newblock Specialization in the Left Prefrontal Cortex for Sentence
  Comprehension.
\newblock Neuron.  2002; 35(3):589--597.
\newblock
  \urlprefix\url{https://www.sciencedirect.com/science/article/pii/S0896627302007882},
  \href{https://doi.org/10.1016/S0896-6273(02)00788-2}{\doiprefix
  \detokenize{https://doi.org/10.1016/S0896-6273(02)00788-2}}.

\bibitem[{Hauk et~al.(2004)Hauk, Olaf and Johnsrude, Ingrid and Pulvermüller,
  Friedemann}]{Hauk2004}
\textbf{\color{eLifeMediumGrey} Hauk O}, Johnsrude I, Pulvermüller F.
\newblock Somatotopic representation of action words in human motor and
  premotor cortex.
\newblock Neuron.  2004; 41(2):301--307.
\newblock
  \urlprefix\url{http://www.sciencedirect.com/science/article/pii/S0896627303008389}.

\bibitem[{de~Heer et~al.(2017)de Heer, Wendy A. and Huth, Alexander G. and
  Griffiths, Thomas L. and Gallant, Jack L. and Theunissen, Frédéric
  E.}]{deheerhierarchical2017}
\textbf{\color{eLifeMediumGrey} de~Heer WA}, Huth AG, Griffiths TL, Gallant JL,
  Theunissen FE.
\newblock The {Hierarchical} {Cortical} {Organization} of {Human} {Speech}
  {Processing}.
\newblock The Journal of Neuroscience.  2017 Jul; 37(27):6539--6557.
\newblock
  \urlprefix\url{http://www.jneurosci.org/lookup/doi/10.1523/JNEUROSCI.3267-16.2017},
  \href{10.1523/JNEUROSCI.3267-16.2017}{\doiprefix
  \detokenize{10.1523/JNEUROSCI.3267-16.2017}}.

\bibitem[{Hewitt and Manning(2019)Hewitt, John and Manning, Christopher
  D.}]{hewitt2019}
\textbf{\color{eLifeMediumGrey} Hewitt J}, Manning CD.
\newblock A {Structural} {Probe} for {Finding} {Syntax} in {Word}
  {Representations}.
\newblock In: \emph{North American Chapter of the Association for Computational
  Linguistics: Human Language Technologies (NAACL)}; 2019. p.~10.
\newblock \urlprefix\url{https://aclanthology.org/N19-1419/}.

\bibitem[{Honnibal and Montani(2017)Honnibal, Matthew and Montani,
  Ines}]{spacy2}
\textbf{\color{eLifeMediumGrey} Honnibal M}, Montani I.
\newblock {spaCy 2}: Natural language understanding with {B}loom embeddings,
  convolutional neural networks and incremental parsing; 2017,
  \urlprefix\url{https://spacy.io/usage}, to appear.

\bibitem[{Hunter(2007)Hunter, John D.}]{matplotlib}
\textbf{\color{eLifeMediumGrey} Hunter JD}.
\newblock Matplotlib: A 2D Graphics Environment.
\newblock Computing in Science \& Engineering.  2007; 9(3):90--95.
\newblock \urlprefix\url{https://ieeexplore.ieee.org/document/4160265},
  \href{10.1109/MCSE.2007.55}{\doiprefix \detokenize{10.1109/MCSE.2007.55}}.

\bibitem[{Huth et~al.(2016)Huth, Alexander G. and de Heer, Wendy A. and
  Griffiths, Thomas L. and Theunissen, Frédéric E. and Gallant, Jack
  L.}]{huthnatural2016}
\textbf{\color{eLifeMediumGrey} Huth AG}, de~Heer WA, Griffiths TL, Theunissen
  FE, Gallant JL.
\newblock Natural speech reveals the semantic maps that tile human cerebral
  cortex.
\newblock Nature.  2016 Apr; 532(7600):453--458.
\newblock \urlprefix\url{http://www.nature.com/articles/nature17637},
  \href{10.1038/nature17637}{\doiprefix \detokenize{10.1038/nature17637}}.

\bibitem[{Jackendoff(2002)Ray Jackendoff}]{Jackendoff2002JACFOL2}
\textbf{\color{eLifeMediumGrey} Jackendoff R}.
\newblock Foundations of Language: Brain, Meaning, Grammar, Evolution.
\newblock Oxford University Press UK; 2002.
\newblock \urlprefix\url{https://academic.oup.com/book/32834}.

\bibitem[{Jain and Huth(2018)Jain, Shailee and Huth,
  Alexander}]{jainincorporating2018}
\textbf{\color{eLifeMediumGrey} Jain S}, Huth A.
\newblock Incorporating Context into Language Encoding Models for fMRI.
\newblock In: Bengio S, Wallach H, Larochelle H, Grauman K, Cesa-Bianchi N,
  Garnett R, editors. \emph{Advances in Neural Information Processing Systems},
  vol.~31 Curran Associates, Inc.; 2018. p.~10.
\newblock
  \urlprefix\url{https://proceedings.neurips.cc/paper/2018/file/f471223d1a1614b58a7dc45c9d01df19-Paper.pdf}.

\bibitem[{Jung-Beeman(2005)Jung-Beeman, Mark}]{jungbeemanbilateral2005}
\textbf{\color{eLifeMediumGrey} Jung-Beeman M}.
\newblock Bilateral brain processes for comprehending natural language.
\newblock Trends in Cognitive Sciences.  2005 Nov; 9(11):512--518.
\newblock
  \urlprefix\url{http://linkinghub.elsevier.com/retrieve/pii/S1364661305002718},
  \href{10.1016/j.tics.2005.09.009}{\doiprefix
  \detokenize{10.1016/j.tics.2005.09.009}}.

\bibitem[{Kinno et~al.(2007)Kinno, Ryuta and Kawamura, Mitsuru and Shioda,
  Seiji and Sakai, Kuniyoshi L.}]{kinnoneural2007}
\textbf{\color{eLifeMediumGrey} Kinno R}, Kawamura M, Shioda S, Sakai KL.
\newblock Neural correlates of noncanonical syntactic processing revealed by a
  pictured sentence matching task.
\newblock Human Brain Mapping.  2007 Oct; 29(9):1015--1027.
\newblock
  \urlprefix\url{https://www.ncbi.nlm.nih.gov/pmc/articles/PMC6871174/},
  \href{10.1002/hbm.20441}{\doiprefix \detokenize{10.1002/hbm.20441}}.

\bibitem[{Kitaev and Klein(2018)Kitaev, Nikita and Klein,
  Dan}]{kitaevklein2018constituency}
\textbf{\color{eLifeMediumGrey} Kitaev N}, Klein D.
\newblock Constituency Parsing with a Self-Attentive Encoder.
\newblock In: \emph{Proceedings of the 56th Annual Meeting of the Association
  for Computational Linguistics (Volume 1: Long Papers)} Melbourne, Australia:
  Association for Computational Linguistics; 2018. p. 2676--2686.
\newblock \urlprefix\url{https://www.aclweb.org/anthology/P18-1249},
  \href{10.18653/v1/P18-1249}{\doiprefix \detokenize{10.18653/v1/P18-1249}}.

\bibitem[{Kundu et~al.(2018)Kundu, Prantik and Voon, Valerie and Balchandani,
  Priti and Lombardo, Michael V. and Poser, Benedikt A. and Bandettini, Peter
  A.}]{kundumultiecho2017}
\textbf{\color{eLifeMediumGrey} Kundu P}, Voon V, Balchandani P, Lombardo MV,
  Poser BA, Bandettini PA.
\newblock Multi-echo {fMRI}: {A} review of applications in {fMRI} denoising and
  analysis of {BOLD} signals.
\newblock NeuroImage.  2018; 154.
\newblock
  \urlprefix\url{http://linkinghub.elsevier.com/retrieve/pii/S1053811917302410},
  \href{10.1016/j.neuroimage.2017.03.033}{\doiprefix
  \detokenize{10.1016/j.neuroimage.2017.03.033}}.

\bibitem[{Lakretz et~al.(2021)Lakretz, Yair and Hupkes, Dieuwke and Vergallito,
  Alessandra and Marelli, Marco and Baroni, Marco and Dehaene,
  Stanislas}]{lakretz2021mechanisms}
\textbf{\color{eLifeMediumGrey} Lakretz Y}, Hupkes D, Vergallito A, Marelli M,
  Baroni M, Dehaene S.
\newblock Mechanisms for handling nested dependencies in neural-network
  language models and humans.
\newblock Cognition.  2021; 213:104699.
\newblock \urlprefix\url{https://arxiv.org/abs/2006.11098}.

\bibitem[{Lakretz et~al.(2019)Lakretz, Yair and Kruszewski, Germ{\'a}n and
  Desbordes, Theo and Hupkes, Dieuwke and Dehaene, Stanislas and Baroni,
  Marco}]{lakretz2019emergence}
\textbf{\color{eLifeMediumGrey} Lakretz Y}, Kruszewski G, Desbordes T, Hupkes
  D, Dehaene S, Baroni M.
\newblock The emergence of number and syntax units in LSTM language models.
\newblock In: \emph{NAACL-HLT (1)}; 2019. p. 11–20.
\newblock \urlprefix\url{https://arxiv.org/abs/1903.07435}.

\bibitem[{LeBel et~al.(2022)LeBel, Amanda and Wagner, Lauren and Jain, Shailee
  and Adhikari-Desai, Aneesh and Gupta, Bhavin and Morgenthal, Allyson and
  Tang, Jerry and Xu, Lixiang and Huth, Alexander G.}]{lebel2022}
\textbf{\color{eLifeMediumGrey} LeBel A}, Wagner L, Jain S, Adhikari-Desai A,
  Gupta B, Morgenthal A, Tang J, Xu L, Huth AG.
\newblock A natural language fmri dataset for Voxelwise Encoding models.
\newblock Biorxiv.  2022;
  \urlprefix\url{https://www.biorxiv.org/content/10.1101/2022.09.22.509104v1},
  \href{10.1101/2022.09.22.509104}{\doiprefix
  \detokenize{10.1101/2022.09.22.509104}}.

\bibitem[{Lerner et~al.(2011)Lerner, Y. and Honey, C. J. and Silbert, L. J. and
  Hasson, U.}]{lernertopographic2011}
\textbf{\color{eLifeMediumGrey} Lerner Y}, Honey CJ, Silbert LJ, Hasson U.
\newblock Topographic {Mapping} of a {Hierarchy} of {Temporal} {Receptive}
  {Windows} {Using} a {Narrated} {Story}.
\newblock Journal of Neuroscience.  2011 Feb; 31(8):2906--2915.
\newblock
  \urlprefix\url{http://www.jneurosci.org/cgi/doi/10.1523/JNEUROSCI.3684-10.2011},
  \href{10.1523/JNEUROSCI.3684-10.2011}{\doiprefix
  \detokenize{10.1523/JNEUROSCI.3684-10.2011}}.

\bibitem[{Li et~al.(2022)Li, Jixing and Shohini Bhattasali and Shulin Zhang and
  Berta Franzluebbers and Wen-Ming Luh and Nathan Spreng and Jonathan R.
  Brennan and Yiming Yang and Christophe Pallier and John Hale}]{LPPdatapaper}
\textbf{\color{eLifeMediumGrey} Li J}, Bhattasali S, Zhang S, Franzluebbers B,
  Luh WM, Spreng N, Brennan JR, Yang Y, Pallier C, Hale J.
\newblock Le Petit Prince Multilingual Naturalistic FMRI Corpus.
\newblock Scientific Data.  2022; 9.
\newblock \urlprefix\url{https://doi.org/10.1038/s41597-022-01625-7}.

\bibitem[{Mar(2011)Mar, Raymond A.}]{mar_neural_2011}
\textbf{\color{eLifeMediumGrey} Mar RA}.
\newblock The neural bases of social cognition and story comprehension.
\newblock Annual review of psychology.  2011; 62:103--134.
\newblock \urlprefix\url{https://pubmed.ncbi.nlm.nih.gov/21126178/}.

\bibitem[{Matchin et~al.(2017)Matchin, William and Hammerly, Christopher and
  Lau, Ellen}]{matchin2017}
\textbf{\color{eLifeMediumGrey} Matchin W}, Hammerly C, Lau E.
\newblock The role of the {IFG} and {pSTS} in syntactic prediction: {Evidence}
  from a parametric study of hierarchical structure in {fMRI}.
\newblock cortex.  2017; 88:106--123.
\newblock Publisher: Elsevier.

\bibitem[{Matchin and Hickok(2020)Matchin, William and Hickok,
  Gregory}]{Matchin2020}
\textbf{\color{eLifeMediumGrey} Matchin W}, Hickok G.
\newblock The Cortical Organization of Syntax.
\newblock Cerebral Cortex.  2020 Mar; 30(3):1481--1498.
\newblock \urlprefix\url{https://pubmed.ncbi.nlm.nih.gov/28088041/},
  \href{10.1093/cercor/bhz180}{\doiprefix \detokenize{10.1093/cercor/bhz180}}.

\bibitem[{Mazoyer et~al.(1993)Mazoyer, B. M. and Tzourio, N. and Frak, V. and
  Syrota, A. and Murayama, N. and Levrier, O. and Salamon, G. and Dehaene, S.
  and Cohen, L. and Mehler, J.}]{mazoyercortical1993}
\textbf{\color{eLifeMediumGrey} Mazoyer BM}, Tzourio N, Frak V, Syrota A,
  Murayama N, Levrier O, Salamon G, Dehaene S, Cohen L, Mehler J.
\newblock The {Cortical} {Representation} of {Speech}.
\newblock Journal of Cognitive Neuroscience.  1993 Oct; 5(4):467--479.
\newblock \urlprefix\url{https://doi.org/10.1162/jocn.1993.5.4.467},
  \href{10.1162/jocn.1993.5.4.467}{\doiprefix
  \detokenize{10.1162/jocn.1993.5.4.467}}, \_eprint:
  https://direct.mit.edu/jocn/article-pdf/5/4/467/1932303/jocn.1993.5.4.467.pdf.

\bibitem[{McKinney et~al.(2010)McKinney, Wes and others}]{mckinney2010data}
\textbf{\color{eLifeMediumGrey} McKinney W}, et~al.
\newblock Data structures for statistical computing in python.
\newblock In: \emph{Proceedings of the 9th Python in Science Conference}, vol.
  445 Austin, TX; 2010. p. 51--56.
\newblock
  \urlprefix\url{https://conference.scipy.org/proceedings/scipy2010/pdfs/mckinney.pdf}.

\bibitem[{Mollica et~al.(2018)Mollica, Francis and Siegelman, Matthew and
  Diachek, Evgeniia and Piantadosi, Steven T. and Mineroff, Zachary and
  Futrell, Richard and Fedorenko, Evelina}]{mollica2018high}
\textbf{\color{eLifeMediumGrey} Mollica F}, Siegelman M, Diachek E, Piantadosi
  ST, Mineroff Z, Futrell R, Fedorenko E.
\newblock High local mutual information drives the response in the human
  language network.
\newblock bioRxiv.  2018; p. 436204.
\newblock
  \urlprefix\url{https://www.biorxiv.org/content/10.1101/436204v1.full}.

\bibitem[{Naselaris et~al.(2011)Naselaris, Thomas and Kay, Kendrick N. and
  Nishimoto, Shinji and Gallant, Jack L.}]{naselarisencoding2011}
\textbf{\color{eLifeMediumGrey} Naselaris T}, Kay KN, Nishimoto S, Gallant JL.
\newblock Encoding and decoding in {fMRI}.
\newblock NeuroImage.  2011 May; 56(2):400--410.
\newblock
  \urlprefix\url{http://linkinghub.elsevier.com/retrieve/pii/S1053811910010657},
  \href{10.1016/j.neuroimage.2010.07.073}{\doiprefix
  \detokenize{10.1016/j.neuroimage.2010.07.073}}.

\bibitem[{Nastase et~al.(2020)Samuel A. Nastase and Ariel Goldstein and Uri
  Hasson}]{nastase2020}
\textbf{\color{eLifeMediumGrey} Nastase SA}, Goldstein A, Hasson U.
\newblock Keep it real: rethinking the primacy of experimental control in
  cognitive neuroscience,.
\newblock NeuroImage.  2020; 222.
\newblock \urlprefix\url{https://www.nature.com/articles/s41597-021-01033-3},
  \href{10.1016/j.neuroimage.2020.117254}{\doiprefix
  \detokenize{10.1016/j.neuroimage.2020.117254}}, publisher: NeuroImage.

\bibitem[{Nastase et~al.(2021)Nastase, Samuel A. and Liu, Yun-Fei and Hillman,
  Hanna and Zadbood, Asieh and Hasenfratz, Liat and Keshavarzian, Neggin and
  Chen, Janice and Honey, Christopher J. and Yeshurun, Yaara and Regev, Mor and
  et al.}]{nastasenarratives}
\textbf{\color{eLifeMediumGrey} Nastase SA}, Liu YF, Hillman H, Zadbood A,
  Hasenfratz L, Keshavarzian N, Chen J, Honey CJ, Yeshurun Y, Regev M, et~al.
\newblock The “narratives” fmri dataset for evaluating models of
  naturalistic language comprehension.
\newblock Scientific Data.  2021; 8(1).
\newblock \href{10.1038/s41597-021-01033-3}{\doiprefix
  \detokenize{10.1038/s41597-021-01033-3}}.

\bibitem[{Newman et~al.(2010)Newman, Sharlene D. and Ikuta, Toshikazu and
  Burns, Thomas}]{newmaneffect2010}
\textbf{\color{eLifeMediumGrey} Newman SD}, Ikuta T, Burns T.
\newblock The effect of semantic relatedness on syntactic analysis: an {fMRI}
  study.
\newblock Brain and language.  2010 May; 113(2):51--58.
\newblock
  \urlprefix\url{https://www.ncbi.nlm.nih.gov/pmc/articles/PMC2854177/},
  \href{10.1016/j.bandl.2010.02.001}{\doiprefix
  \detokenize{10.1016/j.bandl.2010.02.001}}.

\bibitem[{O'Reilly and Frank(2006)O'Reilly, Randall C and Frank, Michael
  J}]{o2006making}
\textbf{\color{eLifeMediumGrey} O'Reilly RC}, Frank MJ.
\newblock Making working memory work: a computational model of learning in the
  prefrontal cortex and basal ganglia.
\newblock Neural computation.  2006; 18(2):283--328.
\newblock \urlprefix\url{https://pubmed.ncbi.nlm.nih.gov/16378516/}.

\bibitem[{Pallier et~al.(2011)Pallier, Christophe and Devauchelle,
  Anne-Dominique and Dehaene, Stanislas}]{palliercortical2011}
\textbf{\color{eLifeMediumGrey} Pallier C}, Devauchelle AD, Dehaene S.
\newblock Cortical representation of the constituent structure of sentences.
\newblock Proceedings of the National Academy of Sciences.  2011;
  108(6):2522--2527.
\newblock \urlprefix\url{https://www.pnas.org/doi/10.1073/pnas.1018711108},
  publisher: National Acad Sciences.

\bibitem[{Pasquiou et~al.(2022)Alexandre Pasquiou and Yair Lakretz and John T.
  Hale and Bertrand Thirion and Christophe Pallier}]{pasquiouicml2022}
\textbf{\color{eLifeMediumGrey} Pasquiou A}, Lakretz Y, Hale JT, Thirion B,
  Pallier C.
\newblock Neural {Language} {Models} are not {Born} {Equal} to {Fit} {Brain}
  {Data}, but {Training} {Helps}.
\newblock In: \emph{Proceedings of the 39th International Conference on Machine
  Learning (ICML)}, vol. 162; 2022. p. 17499--17516.
\newblock \urlprefix\url{https://arxiv.org/abs/2207.03380}.

\bibitem[{Paszke et~al.(2019)Paszke, Adam and Gross, Sam and Massa, Francisco
  and Lerer, Adam and Bradbury, James and Chanan, Gregory and Killeen, Trevor
  and Lin, Zeming and Gimelshein, Natalia and Antiga, Luca and Desmaison, Alban
  and Kopf, Andreas and Yang, Edward and DeVito, Zachary and Raison, Martin and
  Tejani, Alykhan and Chilamkurthy, Sasank and Steiner, Benoit and Fang, Lu and
  Bai, Junjie and Chintala, Soumith}]{NEURIPS20199015}
\textbf{\color{eLifeMediumGrey} Paszke A}, Gross S, Massa F, Lerer A, Bradbury
  J, Chanan G, Killeen T, Lin Z, Gimelshein N, Antiga L, Desmaison A, Kopf A,
  Yang E, DeVito Z, Raison M, Tejani A, Chilamkurthy S, Steiner B, Fang L, Bai
  J, et~al.
\newblock PyTorch: An Imperative Style, High-Performance Deep Learning Library.
\newblock In: \emph{Advances in Neural Information Processing Systems 32}
  Curran Associates, Inc.; 2019.p. 8024--8035.
\newblock
  \urlprefix\url{http://papers.neurips.cc/paper/9015-pytorch-an-imperative-style-high-performance-deep-learning-library.pdf}.

\bibitem[{Pedregosa et~al.(2011)Pedregosa, F. and Varoquaux, G. and Gramfort,
  A. and Michel, V. and Thirion, B. and Grisel, O. and Blondel, M. and
  Prettenhofer, P. and Weiss, R. and Dubourg, V. and Vanderplas, J. and Passos,
  A. and Cournapeau, D. and Brucher, M. and Perrot, M. and Duchesnay,
  E.}]{scikitlearn}
\textbf{\color{eLifeMediumGrey} Pedregosa F}, Varoquaux G, Gramfort A, Michel
  V, Thirion B, Grisel O, Blondel M, Prettenhofer P, Weiss R, Dubourg V,
  Vanderplas J, Passos A, Cournapeau D, Brucher M, Perrot M, Duchesnay E.
\newblock Scikit-learn: Machine Learning in {P}ython.
\newblock Journal of Machine Learning Research.  2011; 12:2825--2830.
\newblock
  \urlprefix\url{https://www.jmlr.org/papers/volume12/pedregosa11a/pedregosa11a.pdf}.

\bibitem[{Pennington et~al.(2014)Pennington, Jeffrey and Socher, Richard and
  Manning, Christopher}]{penningtonglove2014}
\textbf{\color{eLifeMediumGrey} Pennington J}, Socher R, Manning C.
\newblock Glove: {Global} {Vectors} for {Word} {Representation}.
\newblock In: \emph{Proceedings of the 2014 {Conference} on {Empirical}
  {Methods} in {Natural} {Language} {Processing} ({EMNLP})} Doha, Qatar:
  Association for Computational Linguistics; 2014. p. 1532--1543.
\newblock \urlprefix\url{http://aclweb.org/anthology/D14-1162},
  \href{10.3115/v1/D14-1162}{\doiprefix \detokenize{10.3115/v1/D14-1162}}.

\bibitem[{Pereira et~al.(2018{\natexlab{a}})Pereira, Francisco and Lou, Bin and
  Pritchett, Brianna and Ritter, Samuel and Gershman, Samuel J. and Kanwisher,
  Nancy and Botvinick, Matthew and Fedorenko, Evelina}]{pereiratoward2018}
\textbf{\color{eLifeMediumGrey} Pereira F}, Lou B, Pritchett B, Ritter S,
  Gershman SJ, Kanwisher N, Botvinick M, Fedorenko E.
\newblock Toward a universal decoder of linguistic meaning from brain
  activation.
\newblock Nature Communications.  2018 Mar; 9(1):963.
\newblock \urlprefix\url{https://www.nature.com/articles/s41467-018-03068-4},
  \href{10.1038/s41467-018-03068-4}{\doiprefix
  \detokenize{10.1038/s41467-018-03068-4}}, number: 1 Publisher: Nature
  Publishing Group.

\bibitem[{Pereira et~al.(2018{\natexlab{b}})Pereira, Francisco and Lou, Bin and
  Pritchett, Brianna and Ritter, Samuel and Gershman, Samuel J. and Kanwisher,
  Nancy and Botvinick, Matthew and Fedorenko, Evelina}]{pereira_toward_2018}
\textbf{\color{eLifeMediumGrey} Pereira F}, Lou B, Pritchett B, Ritter S,
  Gershman SJ, Kanwisher N, Botvinick M, Fedorenko E.
\newblock Toward a universal decoder of linguistic meaning from brain
  activation.
\newblock Nature Communications.  2018 Mar; 9(1):963.
\newblock \urlprefix\url{http://www.nature.com/articles/s41467-018-03068-4},
  \href{10.1038/s41467-018-03068-4}{\doiprefix
  \detokenize{10.1038/s41467-018-03068-4}}, bandiera\_abtest: a
  Cc\_license\_type: cc\_by Cg\_type: Nature Research Journals Number: 1
  Primary\_atype: Research Publisher: Nature Publishing Group Subject\_term:
  Computational science;Neural decoding Subject\_term\_id:
  computational-science;neural-decoding.

\bibitem[{Pulvermüller(2013)Pulvermüller, Friedemann}]{pulvermuller2013}
\textbf{\color{eLifeMediumGrey} Pulvermüller F}.
\newblock Semantic embodiment, disembodiment or misembodiment? {In} search of
  meaning in modules and neuron circuits.
\newblock Brain and Language.  2013 Oct; 127(1):86--103.
\newblock \href{10.1016/j.bandl.2013.05.015}{\doiprefix
  \detokenize{10.1016/j.bandl.2013.05.015}}.

\bibitem[{Radford et~al.(2019)Radford, Alec and Wu, Jeffrey and Child, Rewon
  and Luan, David and Amodei, Dario and Sutskever, Ilya and
  {others}}]{radfordlanguage2019}
\textbf{\color{eLifeMediumGrey} Radford A}, Wu J, Child R, Luan D, Amodei D,
  Sutskever I, {others}.
\newblock Language models are unsupervised multitask learners.
\newblock OpenAI blog.  2019; 1(8):9.
\newblock
  \urlprefix\url{https://d4mucfpksywv.cloudfront.net/better-language-models/language_models_are_unsupervised_multitask_learners.pdf}.

\bibitem[{Raichle(2015)Raichle, Marcus E.}]{raichle_brains_2015}
\textbf{\color{eLifeMediumGrey} Raichle ME}.
\newblock The brain's default mode network.
\newblock Annual review of neuroscience.  2015; 38:433--447.
\newblock \urlprefix\url{https://pubmed.ncbi.nlm.nih.gov/25938726/}, publisher:
  Annual Reviews.

\bibitem[{Regev et~al.(2013)Regev, M. and Honey, C. J. and Simony, E. and
  Hasson, U.}]{regevselective2013}
\textbf{\color{eLifeMediumGrey} Regev M}, Honey CJ, Simony E, Hasson U.
\newblock Selective and {Invariant} {Neural} {Responses} to {Spoken} and
  {Written} {Narratives}.
\newblock Journal of Neuroscience.  2013 Oct; 33(40):15978--15988.
\newblock
  \urlprefix\url{http://www.jneurosci.org/cgi/doi/10.1523/JNEUROSCI.1580-13.2013},
  \href{10.1523/JNEUROSCI.1580-13.2013}{\doiprefix
  \detokenize{10.1523/JNEUROSCI.1580-13.2013}}.

\bibitem[{Russin et~al.(2019)Russin, Jake and Jo, Jason and O'Reilly, Randall
  C. and Bengio, Yoshua}]{russin2019reilly}
\textbf{\color{eLifeMediumGrey} Russin J}, Jo J, O'Reilly RC, Bengio Y.
\newblock Compositional generalization in a deep seq2seq model by separating
  syntax and semantics; 2019, \urlprefix\url{https://arxiv.org/abs/1904.09708},
  \href{10.48550/ARXIV.1904.09708}{\doiprefix
  \detokenize{10.48550/ARXIV.1904.09708}}, aRXIV.1904.09708.

\bibitem[{Santi and Grodzinsky(2010)Andrea Santi and Yosef
  Grodzinsky}]{SANTI20101285}
\textbf{\color{eLifeMediumGrey} Santi A}, Grodzinsky Y.
\newblock fMRI adaptation dissociates syntactic complexity dimensions.
\newblock NeuroImage.  2010; 51(4):1285--1293.
\newblock
  \urlprefix\url{https://www.sciencedirect.com/science/article/pii/S1053811910003216},
  \href{https://doi.org/10.1016/j.neuroimage.2010.03.034}{\doiprefix
  \detokenize{https://doi.org/10.1016/j.neuroimage.2010.03.034}}.

\bibitem[{Schrimpf et~al.(2020)Schrimpf, Martin and Blank, Idan and Tuckute,
  Greta and Kauf, Carina and Hosseini, Eghbal A. and Kanwisher, Nancy and
  Tenenbaum, Joshua and Fedorenko, Evelina}]{schrimpfartificial2020}
\textbf{\color{eLifeMediumGrey} Schrimpf M}, Blank I, Tuckute G, Kauf C,
  Hosseini EA, Kanwisher N, Tenenbaum J, Fedorenko E.
\newblock Artificial {Neural} {Networks} {Accurately} {Predict} {Language}
  {Processing} in the {Brain}.
\newblock MIT; 2020.

\bibitem[{Shetreet and Friedmann(2014)Shetreet, Einat and Friedmann,
  Naama}]{shetreet2014processing}
\textbf{\color{eLifeMediumGrey} Shetreet E}, Friedmann N.
\newblock The processing of different syntactic structures: fMRI investigation
  of the linguistic distinction between wh-movement and verb movement.
\newblock Journal of Neurolinguistics.  2014; 27(1):1--17.
\newblock
  \urlprefix\url{https://www.sciencedirect.com/science/article/abs/pii/S0911604413000468}.

\bibitem[{Siegelman et~al.(2019)Siegelman, Matthew and Blank, Idan A and
  Mineroff, Zachary and Fedorenko, Evelina}]{siegelman2019attempt}
\textbf{\color{eLifeMediumGrey} Siegelman M}, Blank IA, Mineroff Z, Fedorenko
  E.
\newblock An attempt to conceptually replicate the dissociation between syntax
  and semantics during sentence comprehension.
\newblock Neuroscience.  2019; 413:219--229.
\newblock
  \urlprefix\url{https://www.sciencedirect.com/science/article/pii/S0306452219304026},
  publisher: Elsevier.

\bibitem[{Simony et~al.(2016)Simony, Erez and Honey, Christopher J. and Chen,
  Janice and Lositsky, Olga and Yeshurun, Yaara and Wiesel, Ami and Hasson,
  Uri}]{simony2016}
\textbf{\color{eLifeMediumGrey} Simony E}, Honey CJ, Chen J, Lositsky O,
  Yeshurun Y, Wiesel A, Hasson U.
\newblock Dynamic reconfiguration of the default mode network during narrative
  comprehension.
\newblock Nature Communications.  2016 Jul; 7(1):12141.
\newblock \urlprefix\url{http://www.nature.com/articles/ncomms12141},
  \href{10.1038/ncomms12141}{\doiprefix \detokenize{10.1038/ncomms12141}},
  number: 1 Publisher: Nature Publishing Group.

\bibitem[{Sperry(1961)Sperry, Roger Wolcott}]{sperry1961}
\textbf{\color{eLifeMediumGrey} Sperry RW}.
\newblock Cerebral {Organization} and {Behavior}: {The} split brain behaves in
  many respects like two separate brains, providing new research possibilities.
\newblock Science.  1961; 133(3466):1749--1757.
\newblock \urlprefix\url{https://pubmed.ncbi.nlm.nih.gov/17829720/}, publisher:
  American Association for the Advancement of Science.

\bibitem[{Stromswold et~al.(1996)Karin Stromswold and David Caplan and
  Nathaniel Alpert and Scott Rauch}]{STROMSWOLD1996452}
\textbf{\color{eLifeMediumGrey} Stromswold K}, Caplan D, Alpert N, Rauch S.
\newblock Localization of Syntactic Comprehension by Positron Emission
  Tomography.
\newblock Brain and Language.  1996; 52(3):452--473.
\newblock
  \urlprefix\url{https://www.sciencedirect.com/science/article/pii/S0093934X96900243},
  \href{https://doi.org/10.1006/brln.1996.0024}{\doiprefix
  \detokenize{https://doi.org/10.1006/brln.1996.0024}}.

\bibitem[{Toneva and Wehbe(2019)Toneva, Mariya and Wehbe,
  Leila}]{tonevainterpreting2019}
\textbf{\color{eLifeMediumGrey} Toneva M}, Wehbe L.
\newblock Interpreting and improving natural-language processing (in machines)
  with natural language-processing (in the brain).
\newblock In: Wallach H, Larochelle H, Beygelzimer A, Alché-Buc Fd, Fox E,
  Garnett R, editors. \emph{Advances in {Neural} {Information} {Processing}
  {Systems} 32} Curran Associates, Inc.; 2019.p. 14954--14964.
\newblock
  \urlprefix\url{http://papers.nips.cc/paper/9633-interpreting-and-improving-natural-language-processing-in-machines-with-natural-language-processing-in-the-brain.pdf}.

\bibitem[{Ullman(2004)Ullman, Michael T.}]{ullman2004contributions}
\textbf{\color{eLifeMediumGrey} Ullman MT}.
\newblock Contributions of memory circuits to language: the
  declarative/procedural model.
\newblock Cognition.  2004 May; 92(1):231--270.
\newblock
  \urlprefix\url{https://www.sciencedirect.com/science/article/pii/S0010027703002324}.

\bibitem[{Vigliocco(2000)Gabriella Vigliocco}]{vigliocco2000language}
\textbf{\color{eLifeMediumGrey} Vigliocco G}.
\newblock Language processing: The anatomy of meaning and syntax.
\newblock Current Biology.  2000; 10(2):R78--R80.
\newblock
  \urlprefix\url{https://www.sciencedirect.com/science/article/pii/S0960982200002827},
  \href{https://doi.org/10.1016/S0960-9822(00)00282-7}{\doiprefix
  \detokenize{https://doi.org/10.1016/S0960-9822(00)00282-7}}.

\bibitem[{Virtanen et~al.(2020)Virtanen, Pauli and Gommers, Ralf and Oliphant,
  Travis E. and Haberland, Matt and Reddy, Tyler and Cournapeau, David and
  Burovski, Evgeni and Peterson, Pearu and Weckesser, Warren and Bright,
  Jonathan and van der Walt, St{\'e}fan J. and Brett, Matthew and Wilson,
  Joshua and Millman, K. Jarrod and Mayorov, Nikolay and Nelson, Andrew R. J.
  and Jones, Eric and Kern, Robert and Larson, Eric and Carey, C. J. and Polat,
  {\.{I}}lhan and Feng, Yu and Moore, Eric W. and VanderPlas, Jake and Laxalde,
  Denis and Perktold, Josef and Cimrman, Robert and Henriksen, Ian and
  Quintero, E. A. and Harris, Charles R. and Archibald, Anne M. and Ribeiro,
  Ant{\^o}nio H. and Pedregosa, Fabian and van Mulbregt, Paul and Vijaykumar,
  Aditya and Bardelli, Alessandro Pietro and Rothberg, Alex and Hilboll,
  Andreas and Kloeckner, Andreas and Scopatz, Anthony and Lee, Antony and
  Rokem, Ariel and Woods, C. Nathan and Fulton, Chad and Masson, Charles and
  H{\"a}ggstr{\"o}m, Christian and Fitzgerald, Clark and Nicholson, David A.
  and Hagen, David R. and Pasechnik, Dmitrii V. and Olivetti, Emanuele and
  Martin, Eric and Wieser, Eric and Silva, Fabrice and Lenders, Felix and
  Wilhelm, Florian and Young, G. and Price, Gavin A. and Ingold, Gert-Ludwig
  and Allen, Gregory E. and Lee, Gregory R. and Audren, Herv{\'e} and Probst,
  Irvin and Dietrich, J{\"o}rg P. and Silterra, Jacob and Webber, James T. and
  Slavi{\v{c}}, Janko and Nothman, Joel and Buchner, Johannes and Kulick,
  Johannes and Sch{\"o}nberger, Johannes L. and de Miranda Cardoso, Jos{\'e}
  Vin{\'i}cius and Reimer, Joscha and Harrington, Joseph and Rodr{\'i}guez,
  Juan Luis Cano and Nunez-Iglesias, Juan and Kuczynski, Justin and Tritz,
  Kevin and Thoma, Martin and Newville, Matthew and K{\"u}mmerer, Matthias and
  Bolingbroke, Maximilian and Tartre, Michael and Pak, Mikhail and Smith,
  Nathaniel J. and Nowaczyk, Nikolai and Shebanov, Nikolay and Pavlyk,
  Oleksandr and Brodtkorb, Per A. and Lee, Perry and McGibbon, Robert T. and
  Feldbauer, Roman and Lewis, Sam and Tygier, Sam and Sievert, Scott and Vigna,
  Sebastiano and Peterson, Stefan and More, Surhud and Pudlik, Tadeusz and
  Oshima, Takuya and Pingel, Thomas J. and Robitaille, Thomas P. and Spura,
  Thomas and Jones, Thouis R. and Cera, Tim and Leslie, Tim and Zito, Tiziano
  and Krauss, Tom and Upadhyay, Utkarsh and Halchenko, Yaroslav O. and
  V{\'a}zquez-Baeza, Yoshiki and 1.0 Contributors, SciPy}]{Virtanen2020}
\textbf{\color{eLifeMediumGrey} Virtanen P}, Gommers R, Oliphant TE, Haberland
  M, Reddy T, Cournapeau D, Burovski E, Peterson P, Weckesser W, Bright J,
  van~der Walt SJ, Brett M, Wilson J, Millman KJ, Mayorov N, Nelson ARJ, Jones
  E, Kern R, Larson E, Carey CJ, et~al.
\newblock SciPy 1.0: fundamental algorithms for scientific computing in Python.
\newblock Nature Methods.  2020 Mar; 17(3):261--272.
\newblock \urlprefix\url{https://doi.org/10.1038/s41592-019-0686-2},
  \href{10.1038/s41592-019-0686-2}{\doiprefix
  \detokenize{10.1038/s41592-019-0686-2}}.

\bibitem[{Wehbe et~al.(2014)Wehbe, Leila and Murphy, Brian and Talukdar, Partha
  and Fyshe, Alona and Ramdas, Aaditya and Mitchell,
  Tom}]{wehbesimultaneously2014}
\textbf{\color{eLifeMediumGrey} Wehbe L}, Murphy B, Talukdar P, Fyshe A, Ramdas
  A, Mitchell T.
\newblock Simultaneously uncovering the patterns of brain regions involved in
  different story reading subprocesses.
\newblock PloS one.  2014; 9(11).
\newblock
  \urlprefix\url{https://journals.plos.org/plosone/article?id=10.1371/journal.pone.0112575},
  publisher: Public Library of Science.

\bibitem[{Wolf(2020)Thomas Wolf}]{wolf20}
\textbf{\color{eLifeMediumGrey} Wolf T}.
\newblock Huggingface; 2020, \urlprefix\url{https://huggingface.co/}, to
  appear.

\bibitem[{Xu et~al.(2005)Xu, Jiang and Kemeny, Stefan and Park, Grace and
  Frattali, Carol and Braun, Allen}]{xu_language_2005}
\textbf{\color{eLifeMediumGrey} Xu J}, Kemeny S, Park G, Frattali C, Braun A.
\newblock Language in context: emergent features of word, sentence, and
  narrative comprehension.
\newblock NeuroImage.  2005 Apr; 25(3):1002--1015.
\newblock
  \urlprefix\url{http://www.sciencedirect.com/science/article/pii/S1053811904007748},
  \href{10.1016/j.neuroimage.2004.12.013}{\doiprefix
  \detokenize{10.1016/j.neuroimage.2004.12.013}}.

\end{thebibliography}


\appendix

\begin{appendixbox}
\nolinenumbers

\section{Models training}
\label{Appendix:ModelTraining}
We trained GloVe and GPT-2 on syntactic or semantic features by adapting both vocabulary size and the associated tokenizer.
Table~\ref{Appendix:table:input-examples} provides examples of the features extracted from a short passage.
After feature extraction, a vocabulary listing all possible feature instances is created for each feature type.
A unique id is then associated to each element of the vocabulary.
The tokenizer converts each feature to its unique id.
Finally, the model is fed sequences of ids and learns to perform its task.

\begin{center}

\resizebox{\textwidth}{!}{%
\begin{tabular}[width=\columnwidth]{c  c | c | c | c | c | c | c | c }
  &  & \multicolumn{7}{c}{Input sequence}    \\
\toprule
Integral &   & \multirow{2}{*}{The}       & \multirow{2}{*}{sixth}  & \multirow{2}{*}{planet}     &  \multirow{2}{*}{was}      &  \multirow{2}{*}{ten}      &  \multirow{2}{*}{times}    &  \multirow{2}{*}{larger}   \\
Features &   &    &   &   &    &     &    &   \\
\midrule
\multirow{2}{*}{Syntactic } & Part-of-Speech & DET   & ADJ   & NOUN  &  VERB     &  NOUN      &  NOUN     &  ADJ \\
 & \multirow{2}{*}{Morphology} & \small{Definite=Def|}  & \small{Degree}   & \small{Number} &  \small{Ind|Sing|Past|}     &  \small{Number}      &  \small{Number}     &  \small{Degree} \\
\multirow{3}{*}{Features}  &  & \small{PronType=Art}  &  \small{=Pos} & \small{=Sing} &  \small{Person=3|Fin}     &  \small{=Card}      &  \small{=Plur}  &  \small{=Cmp} \\
 & Number of  & \multirow{2}{*}{1}   & \multirow{2}{*}{1}   & \multirow{2}{*}{2}  &  \multirow{2}{*}{1}    &  \multirow{2}{*}{1}      &  \multirow{2}{*}{2}     &  \multirow{2}{*}{2} \\
 & Closing Nodes &    &    &   &  &    &   &  \\
\midrule
Semantic & Content  & \multirow{2}{*}{-- }       & \multirow{2}{*}{sixth}  & \multirow{2}{*}{planet}     &  \multirow{2}{*}{-- }      &  \multirow{2}{*}{ten}      &  \multirow{2}{*}{times}    &  \multirow{2}{*}{larger}   \\
Features & words  &    &   &   &    &     &    &   \\
\bottomrule
\end{tabular}}
\captionof{table}{Examples of input sequences given to the neural language models when trained on the different feature spaces.}\label{Appendix:table:input-examples}
\end{center}

The Morphology field contains a list of morphological features, with vertical bar (|) as list separator and with underscore to represent the empty list. 
All features represent attribute-value pairs, with an equals sign (=) separating the attribute from the value. 
In addition, features are selected from the universal feature inventory (\url{https://universaldependencies.org/u/feat/index.html}) and are sorted alphabetically by attribute names. 
It is possible that a feature has two or more values for a given word: Case=Acc,Dat. 
In this case, the values are sorted alphabetically.

Note: for display purposes, the morphology attribute values were removed for ‘was', it was originally equal to ‘Mood=Ind|Number=Sing|Person=3|Tense=Past|VerbForm=Fin'.

\section{Context-limited models}
\label{Appendix:ContextLimitedInput}
Using the same original collection of English novels from Project Gutenberg, we trained three GPT-2 models to probe context integration.
More precisely, we restricted the preceding context (size $k=5, 15$ or $45$ tokens) given to the GPT-2 models during training on the "\emph{Integral dataset}".

When training GPT-2 with a limited amount of contextual information, each input sequence contained $k+5$ tokens: a special token at the beginning, $k$ context tokens, the current token for which we retrieve the activations in order to fit fMRI brain data, the token that is predicted by the current token and the 2 special tokens at the end (the last special end-of-sentence token is always preceded by a token encoding a blank space, we omitted it in the following table).

\begin{center}
\resizebox{\textwidth}{!}{%
\begin{tabular}{c | c | c | c | c | c | c | c | c}
{\Large Special}  & \multicolumn{5}{c}{{\Large Context}} & {\Large Current}  & {\Large Predicted}  & {\Large Special } \\
 {\Large token} & \multicolumn{5}{c}{{\Large (size = 5 tokens)}} &  {\Large token} &  {\Large token} & {\Large  token } \\
\toprule
|<endoftext>| & {\LARGE Once}    &  {\LARGE,}    & {\LARGE when}  & {\LARGE I }       &  {\LARGE was }     &  {\LARGE six   }   & {\LARGE years} &  |<endoftext>| \\
\midrule
|<endoftext>| & {\LARGE , }      & {\LARGE when}  & {\LARGE I  }   &  {\LARGE was}      &  {\LARGE six }     &  {\LARGE years}    &  {\LARGE old } &  |<endoftext>| \\
\midrule
|<endoftext>| & {\LARGE when }   &  {\LARGE I }   & {\LARGE was }  &  {\LARGE six}      &  {\LARGE years }   &  {\LARGE old }     &  {\LARGE , }   &  |<endoftext>| \\
\midrule
|<endoftext>| & {\LARGE I  }     &  {\LARGE was}  & {\LARGE six }  &  {\LARGE years }   & {\LARGE old }     &  {\LARGE , }       & {\LARGE I }    &  |<endoftext>| \\
\bottomrule
\end{tabular}}
\captionof{table}{Examples of context-limited input sequences given to GPT-2 for the analyses on context-integration. Here the context size $k$ is equal to 5.}
\end{center}

\clearpage

\section{Removing absolute position information in GPT-2 trained on semantic features}
\label{Appendix:PositionalEmbeddings}

For the GTP-2 model trained on the semantic features, small modifications had to be made to the model architecture in order to remove all residual syntax.
By default, GPT-2 encodes the absolute positions of tokens in sentences.
As word ordering might contain syntactic information, we had to make sure that it could not be leveraged by GPT-2 by means of its positional embeddings, yet keeping information about word proximity as it influences semantics.
We achieved it by slightly modifying the architecture of GPT-2: we first removed the default positional embeddings, and added to the attention scores embeddings encoding relative positions between input tokens.
Indeed, just removing positional embeddings would have led to a bag-of-words model.
By adding these embeddings encoding relative position to the attention scores a token will weight the attention granted to another token depending on their distance. 
By doing so, information about absolute and relative positions is removed from tokens' embeddings as it is not directly added to the tokens' hidden states.
The following explains how this operation was performed.
Let $\textbf{c}_{W}$ = ($c_{w_1}$, . . . , $c_{w_m}$) be a sequence of $m$ tokenized content words. $\textbf{c}_{W}$ is then fed to a $n_{layers}$ transformer with $n_{heads}$ of dimension $d_{heads}$ that first build an embedding representation $\mathbf{E}_{i}, i=1..m$ (of size $d = d_{heads} * n_{heads}$) to which it appends (by default) a position embedding $\textbf{p}_{i}, i=1..m$ (of size $d$) for each token.
To remove all syntactic content, the first step is to discard the previously mentioned positional embeddings $\textbf{p}_{i}, i=1..m$. 
However stopping here would only lead to a bag-of-word model where a given token might be influenced similarly by an adjacent token or one far away.
As a consequence, we had to weight the attention score granted to a token depending on its relative distance.

The attention operation can be described as mapping a query (Q) and a set of key-value (K, V) pairs to an output,
where the query, keys, values, and output are all vectors (generally packed into matrices). The output is computed as a weighted sum of the values, where the weight assigned to each value is computed by a compatibility function of the query with the corresponding key.
We thus modify the classical attention operation:
$$\mbox{Attention}(\textbf{Q}, \textbf{K}, \textbf{V})= \mbox{Softmax}((\textbf{Q} \textbf{K}^{T})/\sqrt{d_{k}})\textbf{V} $$ by adding the previously described relative positional embedding $\textbf{W}$ in the attention mechanisms: $$\mbox{Attention}(\textbf{Q}, \textbf{K}, \textbf{V})= \mbox{Softmax}((\textbf{Q} \textbf{K}^{T} + \textbf{W})/\sqrt{d_{k}})\textbf{V} $$
To build $\textbf{W}$, we first defined the matrix $\textbf{D} = (n -1 + j-i)_{i,j=1..m} \in \mathbb{R}^{m \times m}$ (encoding the number of tokens separating two tokens in the input sequence shifted by $n-1$) for each input sequence $\textbf{c}_{W}$, where $n$ is the maximal input size. 
$\textbf{D}$ is then embedded using a lookup table that stores an embedding of size ($d_{head}$) for each possible value of $\textbf{D}$, giving $\textbf{U}$ ($\in \mathbb{R}^{ m \times m \times d_{head}}$).

Finally, the weights assigned to the value vectors are adjusted using the embedded relative distances between tokens $\textbf{W}$ ($\in \mathbb{R}^{ n_{heads} \times m \times m}$), defined as:

$$W_{i, j, k} = \sum_{d=1}^{d_{head}} K_{i, j, d} U_{j, k, d}$$

By doing so, we were able to weight words interactions depending on their relative distance in the input sequence, while removing all absolute positional information from tokens hidden-states.


\section{Convergence of the language models during training}
\label{Appendix:ModelConvergence}

\begin{center}
\includegraphics[width=\linewidth]{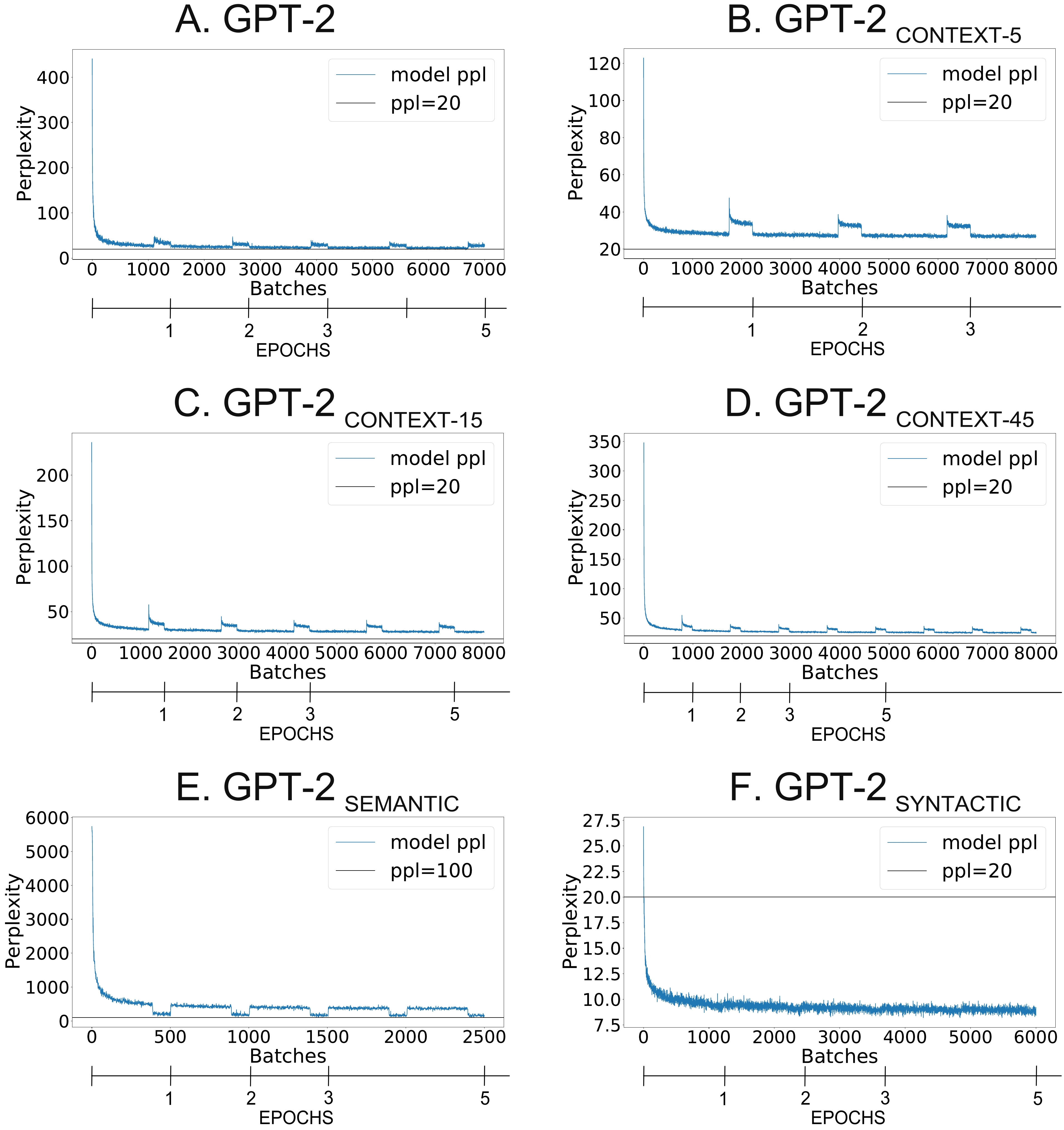}
\captionof{figure}{\textbf{Model convergence during training.} The models represented in panels A to D were trained on the integral features. Models in panels E and F were respectively  trained on the semantic and syntactic features.}
\label{Appendix:Figure:ModelConvergence}
\end{center}


\clearpage

\section{Mapping NLM activations to brain data}
\label{Appendix:MappingNlmToBrain}

Given two non-linear transformations $\varphi_{1}$ (the neural language model that takes as input the sentence and from which we extract latent representations) and $\varphi_{2}$ (the brain that takes as input the sentence and from which we extract voxels' activations) and an input sequence \textbf{w} = (w$_{1}$, . . . , w$_{M}$), we define $\textbf{Y}_{s} = \varphi_{2}(\textbf{w}) \in \mathbb{R}^{N \times V}$ and $\textbf{X} = \varphi_{1}(\textbf{w}) \in \mathbb{R}^{M \times d}$, and we aimed at finding a linear transformation from $\textbf{X}$ to $\textbf{Y}_{s}$, where d is the dimension of the model, V is the number of brain voxels, and N the number of fMRI scans acquired.
One issue is that $\mathbf{X}$ and $\mathbf{Y}_{s}$ don't have the same sampling frequency: $\mathbf{X}$ being defined at word-level while $\mathbf{Y}_{s}$ has been re-sampled at the fMRI acquisition frequency, every 2 seconds.
To map $\mathbf{X}$ to $\mathbf{Y}_{s}$ we first need to temporally align them, taking the dynamic of the fMRI BOLD signal into account, and then determine a linear spatial mapping between the convolved and re-sampled $\mathbf{X}$ and $\mathbf{Y}_{s}$.
Using the standard model-based encoding approach to modelling fMRI signals \citep{naselarisencoding2011,huthnatural2016,pasquiouicml2022}, we first convolve each column of $\mathbf{X}$ with the \textit{SPM} haemodynamic kernel (K), which corresponds to the profile of the fMRI BOLD response following a Dirac stimulation, and then sub-sampled the signal to match the sampling frequency of $\mathbf{Y}_{s}$, giving $\tilde{\mathbf{X}} = S_{ub}(K \circ X)$, with $S_{ub}$ the sub-sampling operator.
Finally, we learn the linear spatial mapping between $\tilde{\mathbf{X}}$ and $\mathbf{Y}_{s}$ using a nested cross-validated L2-regularized (aka Ridge) univariate linear encoding model.
More precisely, for each voxel $\mathbf{y}_{s}^{v} $, we learn a linear projection 
$\bm{\hat{\beta}}_{s}^{v}$ 
from $\tilde{\mathbf{X}}$ to $\mathbf{y}_{s}^{v} $ using a nested cross-validated L2-regularized univariate linear encoding model whose general solution is given by:
$$
\bm{\hat{\beta}_s^v} = arg\min_{\bm{\beta_s}}\|\mathbf{y}_s^v-\bm{\beta_s}^T \mathbf{X} \|^2 + \lambda\|\bm{\beta_s}\|_{2}^{2}
\; \mbox{i.e.} \; \bm{\hat{\beta}}_{s} = \mbox{Ridge}(\mathbf{X}, \bm{Y_{s}}) 
$$
The latter stage resulted for each model and each run into a design matrix $\mathbf{X}$ of size $N \times d$. Given a neural language model, we gave the associated nine design-matrices to a nested cross-validated L2-regularized univariate linear encoding model to fit the fMRI brain data (of size $N \times V$). To evaluate model performance and the optimal regularization parameter $\lambda^*$, we used a nested cross-validation procedure: we split each participant's dataset into training, validation and test sets, such that the training set included 7 out of the 9 experiment runs, and the validation and test sets contained one of the two remaining sessions. We evaluated model performance using Pearson correlation coefficient $R$, which is a measure of the linear correlation between encoding models' predicted time-courses and the actual time-courses.
For each subject and each voxel, we first determined  $\lambda^*$ by comparing $R_{valid}$ for 10 different values of $\lambda$, linearly spaced in log-scale between $10^{-3}$ and $10^4$. We then calculated $R_{test}$ for  $\lambda^*$. Finally, we repeated this procedure 9 times, using cross-validation. This resulted in 9 $R_{test}$ values that we then averaged to produce a single $R_{test}$ map for the participant.
We evaluated the quality of the mapping for subject $s$ in voxel $v$ using Pearson correlation:
$$
R(X)_{s}^v = \mbox{Corr}(\bf{Y_{s}^v}, \bm{\hat{\beta}_s^v} \bf{X})
$$

\clearpage

\section{The Basic Features baseline model}
\label{Appendix:BF}

To assess the specific impact of NLMs' embeddings, the maps shown in Fig.\ref{Figure:BrainFit} report \emph{increases in R values} relative to a \emph{baseline model} which comprised three variables of non-interest:
\begin{itemize}
\item acoustic energy (root mean squared of the audio signal sampled every 10ms)
\item word offsets (one event at each word offset)
\item log of the lexical frequency of each word (modulator of the words events).
\end{itemize}

More generally, as we looked at increases in $R$ scores between models, the baseline model was appended to all other models studied in order to cancel out the effects of the 3 features of non-interest.
Appendix 1-Fig.\ref{Appendix:Figure:BF} below displays the cross-validated correlations obtained from this baseline model.

\begin{center}
\includegraphics[width=0.8\columnwidth]{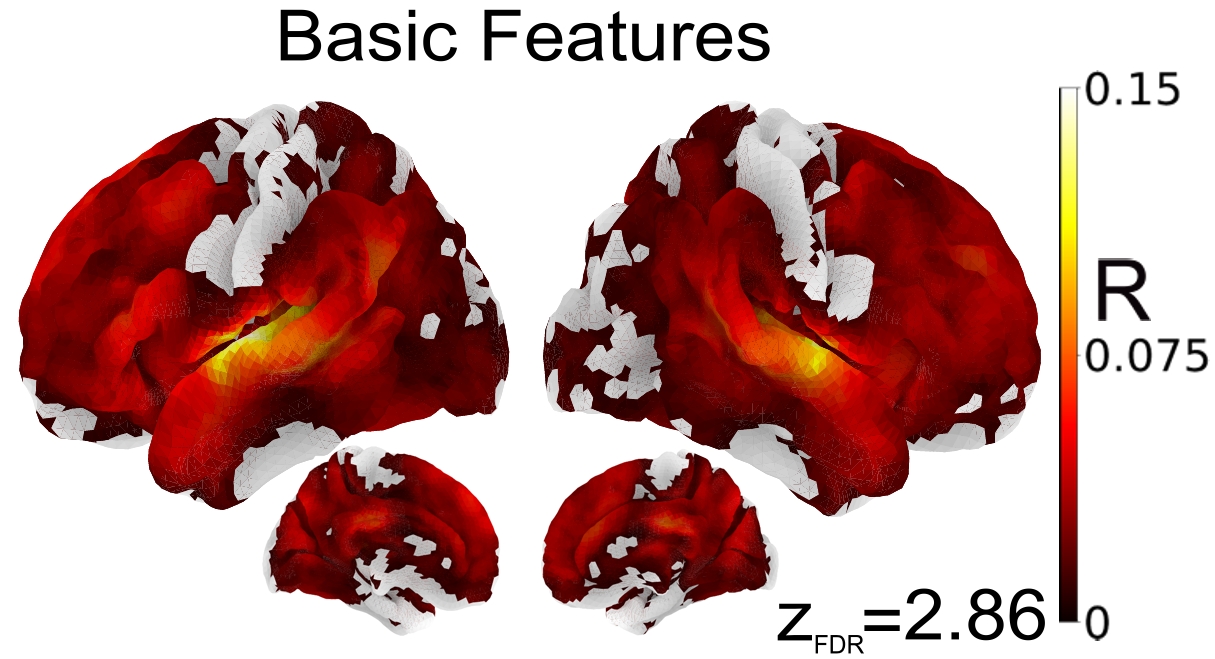}
\captionof{figure}{\textbf{Brain regions showing significant activations for the Basic Features baseline model.} Using the Basic Features (BF) baseline model to fit fMRI brain data, we displayed voxels where there was a significant correlation (voxel-wise thresholded group analyses; N=51 subjects; corrected for multiple
comparisons with a FDR approach $p < 0.005$; $z_{FDR}$ is the FDR threshold on the z-scores). The effects from the Basic Features baseline model were discarded from all the analyses in the paper.
}
\label{Appendix:Figure:BF}
\end{center}


\section{Brain fit of GloVe and GPT-2 when trained on the Integral Features}
\label{Appendix:BrainFit}

Appendix 1-Fig.\ref{Appendix:Figure:BrainFit} shows the increase in $R$, relative to the baseline model, provided by the GloVe and GPT-2 models trained on the Integral Features, that is, the intact text. 

\begin{center}
\includegraphics[width=\columnwidth]{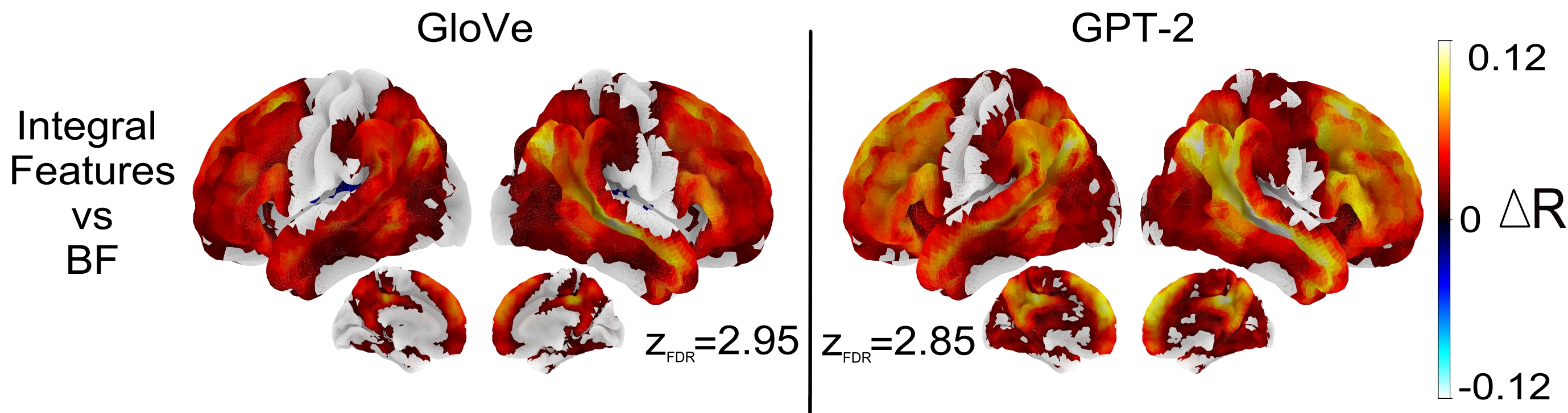}
\captionof{figure}{\textbf{Brain regions showing significant $R$ score increases compared to the Baseline Model for GloVe and GPT-2 when trained on the Integral Features.} Increases in R scores relative to the baseline model for GloVe (a non contextual model) and GPT-2 (a contextual model), trained on the Integral features (voxel-wise thresholded group analyses; N=51 subjects; corrected for multiple comparisons with a FDR approach $p < 0.005$; $z_{FDR}$ is the FDR threshold on the z-scores).}
\label{Appendix:Figure:BrainFit}
\end{center}


\section{R Scores Distribution for GloVe and GPT-2 Trained on Semantic or Syntactic Features}
\label{Appendix:RScoreDistribution}

Appendix 1-Fig.\ref{Appendix:Figure:RScoreDistribution} below shows the averaged (across participants) voxels distribution, of the increase in $R$ scores obtained from GloVe and GPT-2 models on semantic or syntactic features, relative to the baseline model.

\begin{center}
\includegraphics[width=\columnwidth]{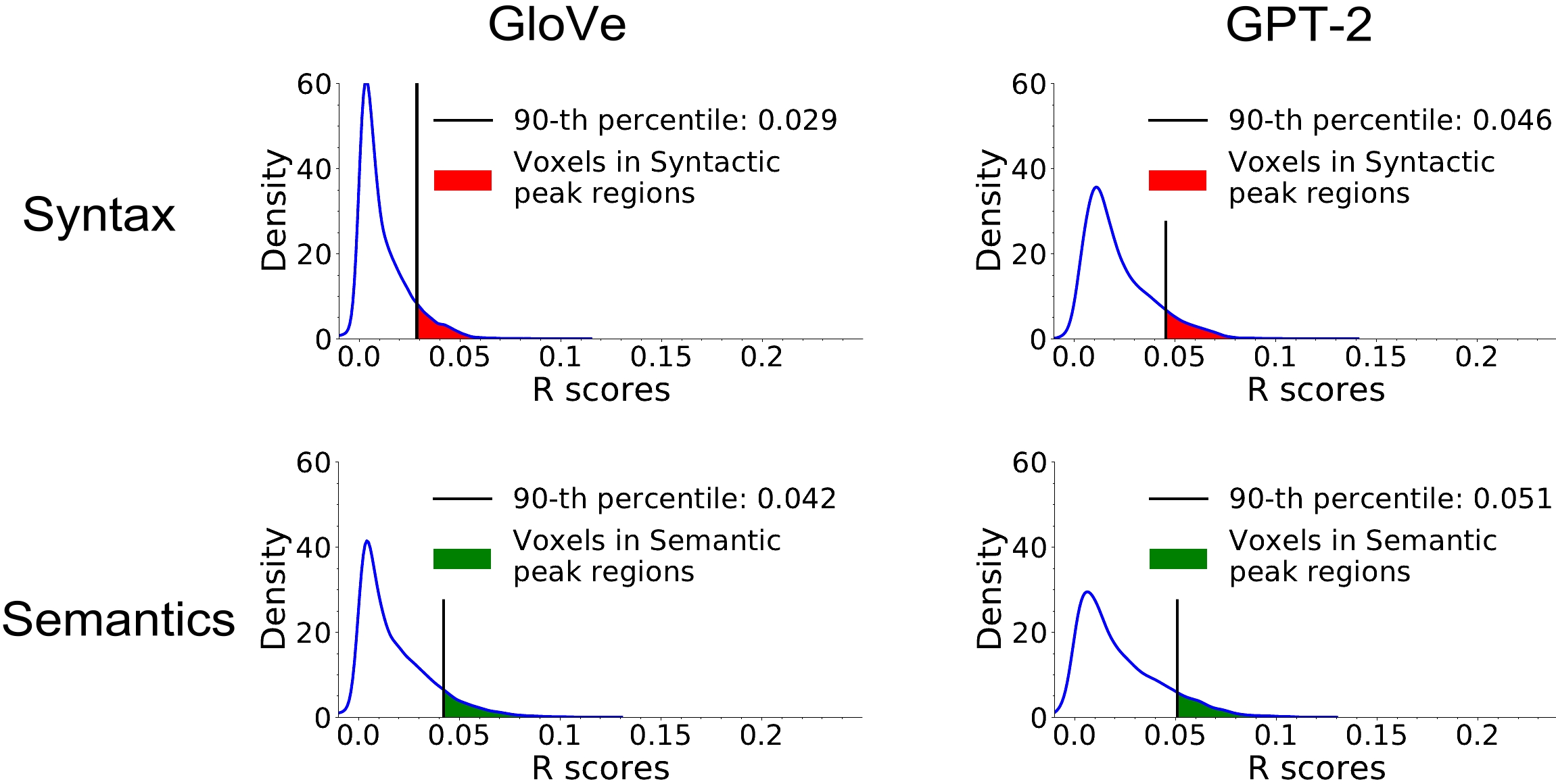} 
\centering
\captionof{figure}{\textbf{Distribution of $R$ scores derived from GloVe and GPT-2 semantic and syntactic embeddings.} The 90th-percentile of the $R$ scores distribution is highlighted with a vertical black line and used to select voxels for the peak regions analyses.
}
  
\label{Appendix:Figure:RScoreDistribution}

\end{center}

\clearpage

\section{Comparison of the models trained on Semantic features with the models trained on Syntactic features}
\label{Appendix:DiffRatio}

Appendix 1-Fig.\ref{Appendix:Figure:DiffRatio} shows the differences in R scores between the semantic and syntactic models, for Glove and GPT-2. 
Correcting for multiple comparisons (N=51; $p < 0.005$ after FDR correction), we observed significant differences in favor of the syntactic embeddings in the STG, and significant differences in favor of the semantic embeddings in the pMTG, the AG and the IFS and SFS.

\begin{center}
\includegraphics[width=\columnwidth]{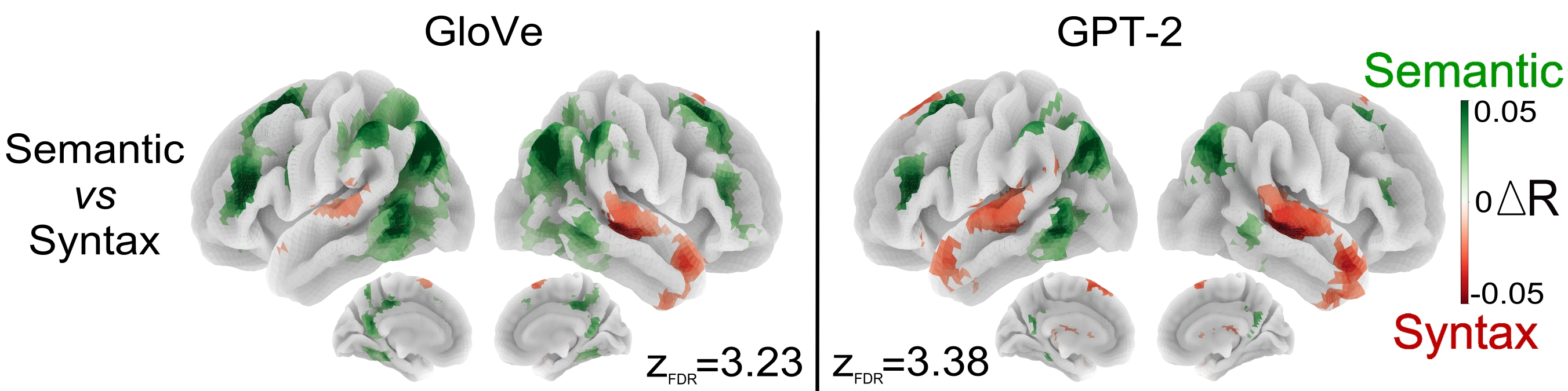}
\captionof{figure}{\textbf{Comparison of the models trained on Semantic features with the models trained on Syntactic features.} Significant R score differences between the models trained on Semantic features and the models trained on Syntactic features. The brain regions that are better fitted by the former model appear in green, while the regions better fitted by the latter model appear in red. (All these maps represent voxel-wise thresholded group analyses; N=51 subjects; corrected for multiple comparisons with a FDR approach $p<0.005$).
}
\label{Appendix:Figure:DiffRatio}
\end{center}

\clearpage

\section*{Brain Regions abbreviations}
\label{Appendix:SpecificityIndex}

\begin{itemize}
    \item STG: superior Temporal Gyrus
    \item STS: superior Temporal Sulcus
    \item TP: Temporal Pole
    \item IFG: inferior Frontal Gyrus
    \item IFS: inferior Frontal Sulcus
    \item DMPC:  Dorso-Medial Prefrontal Cortex
    \item pMTG: posterior Middel Temporal Gyrus
    \item TPJ: temporo-parietal junction
    \item pCC: posterior Cingulate Cortex
    \item AG: Angular Gyrus
    \item SMA: Supplementary Motor Area
\end{itemize}

\clearpage

\end{appendixbox}

\end{document}